\documentclass[lettersize,journal]{IEEEtran}
\usepackage{amsmath,amsfonts}
\usepackage{algorithmic}
\usepackage{algorithm}
\usepackage{array}
\usepackage[caption=false,font=small,labelfont=rm,textfont=rm]{subfig}
\usepackage{textcomp}
\usepackage{stfloats}
\usepackage{url}
\usepackage{verbatim}
\usepackage{graphicx}
\usepackage{cite}
\usepackage{ragged2e} 
\usepackage{booktabs,makecell,multirow,tabularx}
\usepackage[colorlinks,
            linkcolor=blue,
            anchorcolor=blue,
            citecolor=blue]{hyperref}
\usepackage[table]{xcolor}
\begin{document}

\title{Efficient Feature Aggregation and Scale-Aware Regression for Monocular 3D Object Detection}

\author{Yifan Wang, 
        Xiaochen Yang, 
        Fanqi Pu, \\
        Qingmin Liao,~\IEEEmembership{Senior Member,~IEEE,}
        and Wenming Yang,~\IEEEmembership{Senior Member,~IEEE}
\thanks{Yifan Wang, Fanqi Pu, Qingmin Liao, and Wenming Yang are with the Department of Electronic Engineering, Shenzhen International Graduate School, Tsinghua University, Shenzhen 518055, China
(e-mail: yf-wang23@mails.tsinghua.edu.cn; pfq23@mails.tsinghua.edu.cn; liaoqm@tsinghua.edu.cn; yang.wenming@sz.tsinghua.edu.cn).}

\thanks{Xiaochen Yang is with the School of Mathematics and Statistics, University of Glasgow, Glasgow, G12 8QQ, UK
(e-mail: xiaochen.yang@glasgow.ac.uk).}
}

\markboth{Journal of \LaTeX\ Class Files,~Vol.~14, No.~8, August~2021}%
{Shell \MakeLowercase{\textit{et al.}}: }


\maketitle

\begin{abstract}
Monocular 3D object detection has received considerable attention for its simplicity and low cost.
Existing methods typically follow conventional 2D detection paradigms, first locating object centers and then predicting 3D attributes via neighboring features. 
However, these approaches mainly focus on local information, which may limit the model's global context awareness and result in missed detections, as the global context provides semantic and spatial dependencies essential for detecting small objects in cluttered or occluded environments.
In addition, due to large variation in object scales across different scenes and depths, inaccurate receptive fields often lead to background noise and degraded feature representation.
To address these issues, we introduce MonoASRH, a novel monocular 3D detection framework composed of Efficient Hybrid Feature Aggregation Module (EH-FAM) and Adaptive Scale-Aware 3D Regression Head (ASRH).
Specifically, EH-FAM employs multi-head attention with a global receptive field to extract semantic features and leverages lightweight convolutional modules to efficiently aggregate visual features across different scales, enhancing small-scale object detection.
The ASRH encodes 2D bounding box dimensions and then fuses scale features with the semantic features aggregated by EH-FAM through a scale-semantic feature fusion module. 
The scale-semantic feature fusion module guides ASRH in learning dynamic receptive field offsets, incorporating scale information into 3D position prediction for better scale-awareness. 
Extensive experiments on the KITTI and Waymo datasets demonstrate that MonoASRH achieves state-of-the-art performance. The code and model are released at \url{https://github.com/WYFDUT/MonoASRH}.
\end{abstract}

\begin{IEEEkeywords}
3D object detection, monocular, scale-aware, scene understanding, autonomous driving.
\end{IEEEkeywords}

\section{Introduction}
\IEEEPARstart{I}{n recent} years, 3D object detection has emerged as a pivotal area of research, particularly driven by its critical role in autonomous systems, robotics, and augmented reality. The ability to accurately detect and locate objects in three-dimensional space is essential for applications that require a high degree of spatial awareness, especially for autonomous driving systems\cite{arnold2019survey},\cite{grigorescu2020survey}.

Advancements in sensors such as LiDAR, radar, and stereo cameras have enhanced 3D object detection with point cloud data, radar signals, and depth maps. However, LiDAR's high-resolution point clouds\cite{chen2021monorun},\cite{reading2021categorical},\cite{huang2022monodtr},\cite{chong2022monodistill} and stereo methods' depth maps\cite{li2020confidence},\cite{zhou2022sgm3d} require additional calibration. CAD\cite{liu2021autoshape},\cite{li2022densely} and multi-frame\cite{brazil2020kinematic} approaches improve shape and movement tracking, but are also computationally intensive. 
Compared to the above methods, monocular vision systems offer a more cost-effective and deployable solution. However, they cannot accurately recover 3D information from 2D images without explicit depth perception.

\begin{figure}
    \centering
    \includegraphics[width=0.95\linewidth]{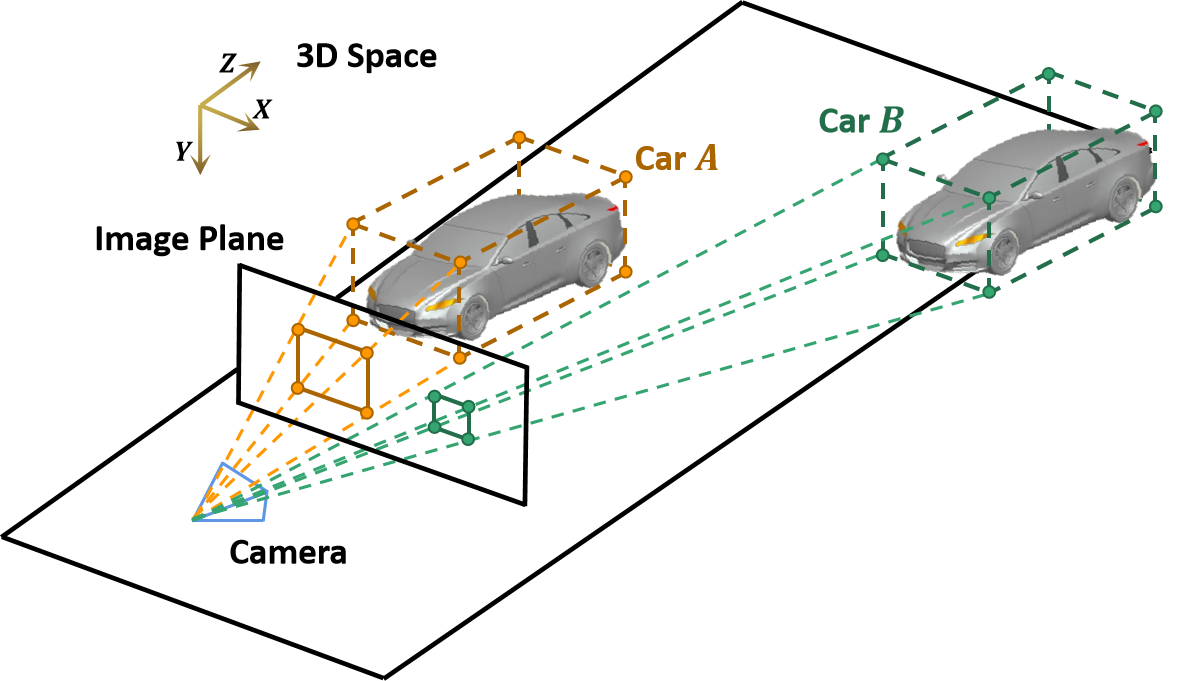}%
    \caption{Schematic representation of vehicle imaging size at varying distances from the camera. Cars farther from the camera tend to occupy a smaller proportion of the image, while those closer tend to occupy a larger proportion.}
    \label{fig:ditance}
    \vspace{-0.5cm}
\end{figure}

\begin{figure*}[htbp]
    \centering
    \subfloat[ DEVIANT]{
        \centering
        \includegraphics[width=0.3075\linewidth]{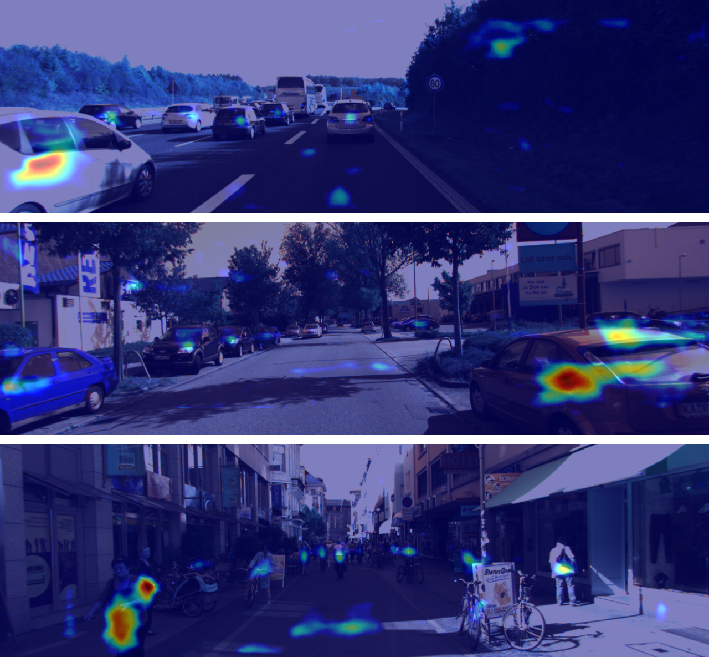}%
        \label{fig:heatmapa}%
    }\hfill
    \subfloat[ MonoLSS]{
        \centering
        \includegraphics[width=0.3075\linewidth]{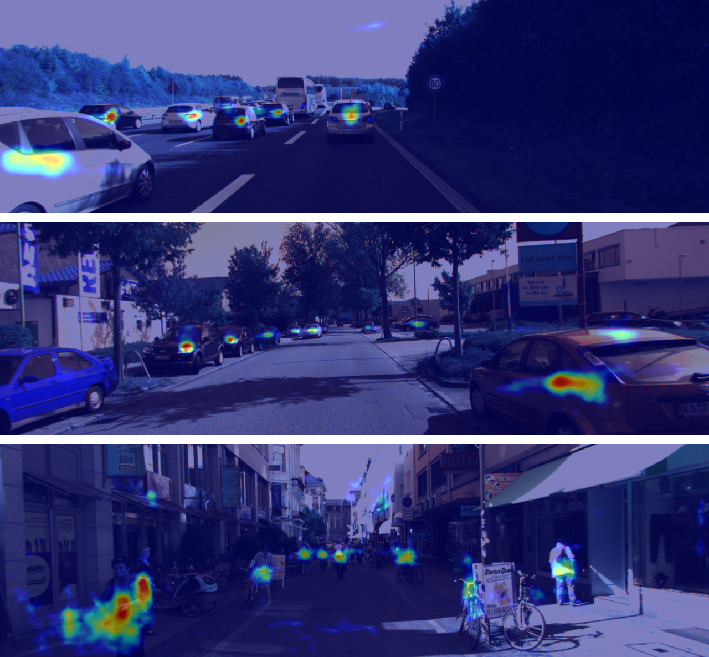}%
        \label{fig:heatmapb}%
    }\hfill
    \subfloat[ MonoASRH]{
        \centering
        \includegraphics[width=0.3075\linewidth]{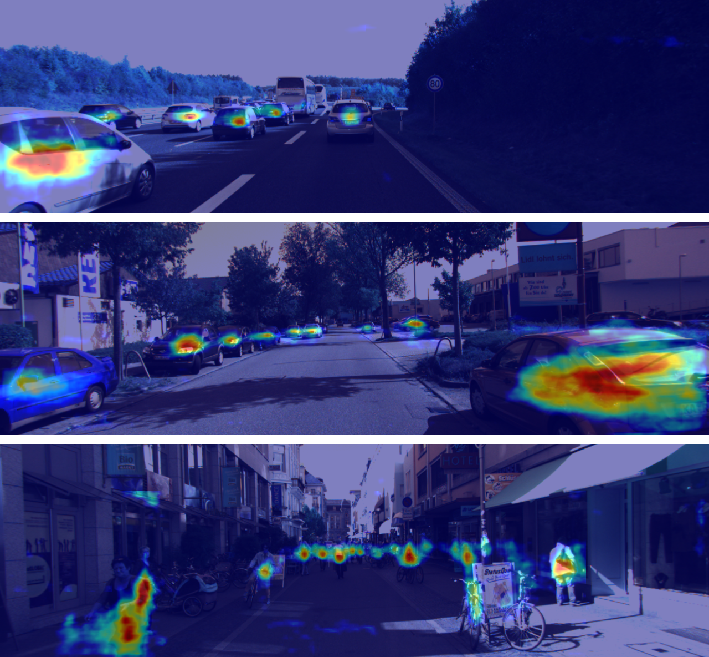}%
        \label{fig:heatmapc}%
    }\hfill
    \subfloat{
        \centering
        \includegraphics[width=0.050\linewidth]{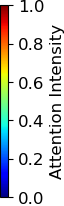}%
        \label{fig:colorbar}%
    }
    \caption{Visualization of attention heatmaps for different models: (a) DEVIANT\cite{kumar2022deviant}, (b) MonoLSS\cite{li2024monolss}, and (c) Our proposed MonoASRH. Previous methods struggle to capture distant and occluded objects due to fixed receptive fields. In contrast, MonoASRH dynamically adjusts attention across multiple scales, improving detection of various object categories, such as cars and pedestrians.}
    \label{fig:heatmap}
    \vspace{-0.5cm}
\end{figure*}

Therefore, current advances in monocular 3D detection have focused on improving depth estimation accuracy. Methods like\cite{zhou2019objects},\cite{liu2020smoke},\cite{ma2021delving},\cite{luo2021m3dssd},\cite{liu2022learning},\cite{li2024monolss} following the CenterNet\cite{duan2019centernet} paradigm, use direct regression to estimate the 3D center depth of the target object.\cite{liu2021autoshape},\cite{li2020rtm3d},\cite{zhang2021objects},\cite{lu2021geometry},\cite{kumar2022deviant} incorporate geometric constraints into depth estimation to enhance target depth recovery.\cite{kim2022boosting},\cite{li2022diversity},\cite{qin2022monoground},\cite{zhu2023monoedge},\cite{yan2024monocd} effectively integrate geometric depth with regressed depth, improving the network's generalization ability. 
However, these approaches focus on decoupling the regression of 2D and 3D attributes: predicting 2D bounding box parameters $\left\{ {{x_{2D}},{y_{2D}},{w_{2D}},{h_{2D}}} \right\}$ and 3D object parameters $\left\{ {{x_{3D}},{y_{3D}},{z_{3D}},{w_{3D}},{h_{3D}},{l_{3D}},yaw} \right\}$ separately. This separation may neglect valuable spatial cues provided by the 2D predictions, such as the object’s size and position in the image, which can serve as geometric information for estimating depth and 3D structure. For instance, distant objects tend to produce smaller 2D bounding boxes, while nearby objects appear larger (Fig. \ref{fig:ditance}), indicating a potential correlation between 2D scale and 3D distance that is not fully exploited in decoupled frameworks.
This issue is evident in the contrast between the attention heatmaps of the networks shown in Fig. \ref{fig:heatmap}. 
It can be seen that DEVIANT\cite{kumar2022deviant} and MonoLSS\cite{li2024monolss} struggle to focus on smaller, distant objects, particularly in the pedestrian category. Moreover, DEVIANT\cite{kumar2022deviant} tends to inadvertently take care of irrelevant background noise.
Some previous works attempted to address this issue. Chen et al. proposed a shape-aware auxiliary training task\cite{chen2023shape}. \cite{chen2023monocular} integrated deformable convolutions\cite{zhu2019deformable} to enhance the adaptability of the model to the features. Although these methods achieve dynamic receptive field adjustment to some extent, they have not explicitly considered object scale within a scene, nor dynamically adjusted the model's attention based on different scales.

To address the aforementioned limitations, this paper proposes the novel \textbf{E}fficient \textbf{H}ybrid \textbf{F}eature \textbf{A}ggregation \textbf{M}odule (EH-FAM) and \textbf{A}daptive \textbf{S}cale-Aware 3D \textbf{R}egression \textbf{H}ead (ASRH). 
Inspired by RT-DETR's\cite{zhao2024detrs} success in 2D detection, EH-FAM integrates Vision Transformers with CNNs to efficiently aggregate visual features, significantly reducing model complexity compared to traditional methods like DLAUp\cite{yu2018deep}. 
The EH-FAM employs self-attention mechanisms for within-scale feature interaction at the highest level, while effectively propagating low-level feature information across all scales through a dedicated CNN architecture.
Subsequently, ASRH encodes 2D bounding box dimensions to capture scale features, which are fused with semantic features extracted from the cross-scale feature through a specially designed scale-semantic feature fusion module.
The fused features guide the learning of receptive field offsets, which are then applied to deformable convolutions, enhancing the model's ability to adapt to varying object scales. 
Furthermore, ASRH utilizes a spatial variance-based attention mechanism to adaptively differentiate foreground objects from noise within semantic features. 
Additionally, we introduce the Selective Confidence-Guided Heatmap Loss to facilitate ASRH to prioritize high confidence detections and mitigate the impact of hard samples.

Our contributions can be summarized as follows:

\begin{itemize}
\item{We introduce a plug-and-play module called Efficient Hybrid Feature Aggregation Module, which is designed for efficient cross-scale feature aggregation.}
\item{We propose a novel Adaptive Scale-Aware 3D Regression Head, which dynamically adjusts the network’s receptive field according to object scale. By leveraging 2D predictions, ASRH facilitates 3D bounding box regression.}
\item{To enhance training stability, we introduce a Selective Confidence-Guided Heatmap Loss that emphasizes high-confidence detections while mitigating the impact of hard samples.}
\item{Extensive experiments on the KITTI 3D object detection benchmark\cite{geiger2012we} and Waymo Open dataset\cite{sun2020scalability} show that our MonoASRH achieves better results compared to previous state-of-the-art methods.}
\end{itemize}

\section{Related Works}

\subsection{Monocular 3D Object Detection}
In general, Monocular 3D object detection methods can be categorized into three main approaches: depth map-based, center-based, and transformer-based methods. Depth map-based approaches address spatial ambiguity by estimating depth maps using supervised \cite{fu2018deep},\cite{lee2019big} or self-supervised networks\cite{godard2019digging}, \cite{guizilini20203d}.\cite{ding2020learning}, with techniques like pseudo-LiDAR conversions\cite{weng2019monocular},\cite{ma2019accurate} and multi-scale depth fusion\cite{wang2021depth},\cite{ma2020rethinking} improving geometric reasoning. However, their reliance on pre-trained depth networks and overlap between training and validation datasets\cite{simonelli2021we} often lead to foreground depth inaccuracies.

Center-based methods extend anchor-free frameworks like CenterNet \cite{duan2019centernet}, linking 3D attributes to object centers. To mitigate depth errors, works like\cite{ma2021delving},\cite{zhang2021objects},\cite{kumar2022deviant} integrate geometric constraints and hierarchical learning, while\cite{li2022diversity} leverages multiple depth attributes to generate various depth estimates. \cite{chen2023shape},\cite{yao2023occlusion} propose shape-aware schemes and plane-constrained 3D detection frameworks, respectively. MonoCD\cite{yan2024monocd} introduces the concept of complementary depth. MonoLSS\cite{li2024monolss} uses Gumbel-Softmax probabilistic sampling\cite{jang2016categorical} to distinguish between positive and negative samples in depth maps. Despite efficiency, these methods risk overlooking 2D-3D geometric consistency. 

Transformer-based models leverage end-to-end architectures\cite{wang2022detr3d},\cite{huang2022monodtr} for global feature interaction, injecting global depth information into the Transformer to guide detection. MonoDETR\cite{zhang2023monodetr} interacts with depth and visual features through a depth cross-attention layer in the decoder. To improve inference efficiency, MonoATT\cite{zhou2023monoatt} introduces an adaptive token Transformer. However, their computational complexity, sensitivity to background noise, and slow convergence hinder real-time applicability.

In this work, we propose a center-based framework called MonoASRH that integrates lightweight convolutional modules with multi-head attention for feature aggregation. To further enhance 3D detection robustness, we introduce a latent space projection paradigm where 2D scale features are mapped to guide the 3D regression heads.

\subsection{Feature Aggregation Strategy}
Predominant center-based monocular 3D detection frameworks\cite{ma2021delving},\cite{zhang2021objects},\cite{kumar2022deviant},\cite{yan2024monocd},\cite{li2024monolss} utilize DLAUp\cite{yu2018deep} for feature aggregation. However, DLAUp's hierarchical architecture relies solely on convolutional operations, limiting its capacity for global context integration. Additionally, its dependence on large-kernel transposed convolutions for upsampling significantly increases parametric complexity. Although MonoDETR\cite{zhang2023monodetr} introduces an end-to-end Transformer architecture to address these limitations, the conventional Transformer Encoder struggles with sensitivity to irrelevant background features and exhibits slow convergence, ultimately constraining detection performance.

Recent research has increasingly focused on hybrid CNN-Transformer architectures for feature aggregation. Approaches such as the U-shaped hybrid network\cite{deng2023cformer}, cross-scale token attention mechanisms\cite{yoo2023enriched}, dual-resolution self-attention designs\cite{ilyas2024hybrid}, and RT-DETR's efficient hybrid encoder\cite{zhao2024detrs} collectively demonstrate the capability of such architectures to model long-range dependencies while retaining localized feature extraction. This hybrid architecture can effectively mitigate the aforementioned challenges in feature aggregation modules of monocular 3D detection methods.

Inspired by RT-DETR, our proposed EH-FAM architecture incorporates a multi-head attention mechanism to process highest-level feature maps. This design captures long-range dependencies within abstract semantic representations, enabling comprehensive global context modeling. Furthermore, we design a lightweight CNN-based cross-scale feature fusion structure that significantly enhances inference efficiency through re-parameterization\cite{ding2021repvgg} and convolutional decoupling\cite{szegedy2017inception} techniques.

\subsection{Scale-Aware Object Detection}
Scale-aware methodologies improve detection robustness by explicitly incorporating dimensional information. SAFR-CNN\cite{li2017scale} introduces scale-aware weighting to prioritize sub-networks specialized for input scales. Extending this paradigm, SFDet\cite{zhang2019single} adapts the strategy for face detection through systematic association of multi-scale feature layers. Complementary research by Chen et al.\cite{chen2021scale} develops Scale-aware method to optimize augmentation policies, whereas AdaZoom\cite{xu2022adazoom} introduces adaptive region zooming conditioned on object scale characteristics. SMT\cite{lin2023scale} further advances this field with Scale-Aware Modulation, integrating hybrid convolutions with lightweight aggregation for context-aware token refinement.

Currently, most methods focus on scale awareness in 2D object detection, whereas there is relatively little work on monocular 3D detection. Notably, scale information constitutes a critical geometric prior for 3D detectors due to the inherent relationship between an object's projected 2D dimensions and its spatial position in 3D space. To address this limitation, our ASRH module encodes 2D scale cues to generate dynamic receptive field offsets that guide the regression head toward scale-aware 3D localization.

\section{Proposed Methodology}
\begin{figure*}
    \centering
    \includegraphics[width=0.90\linewidth]{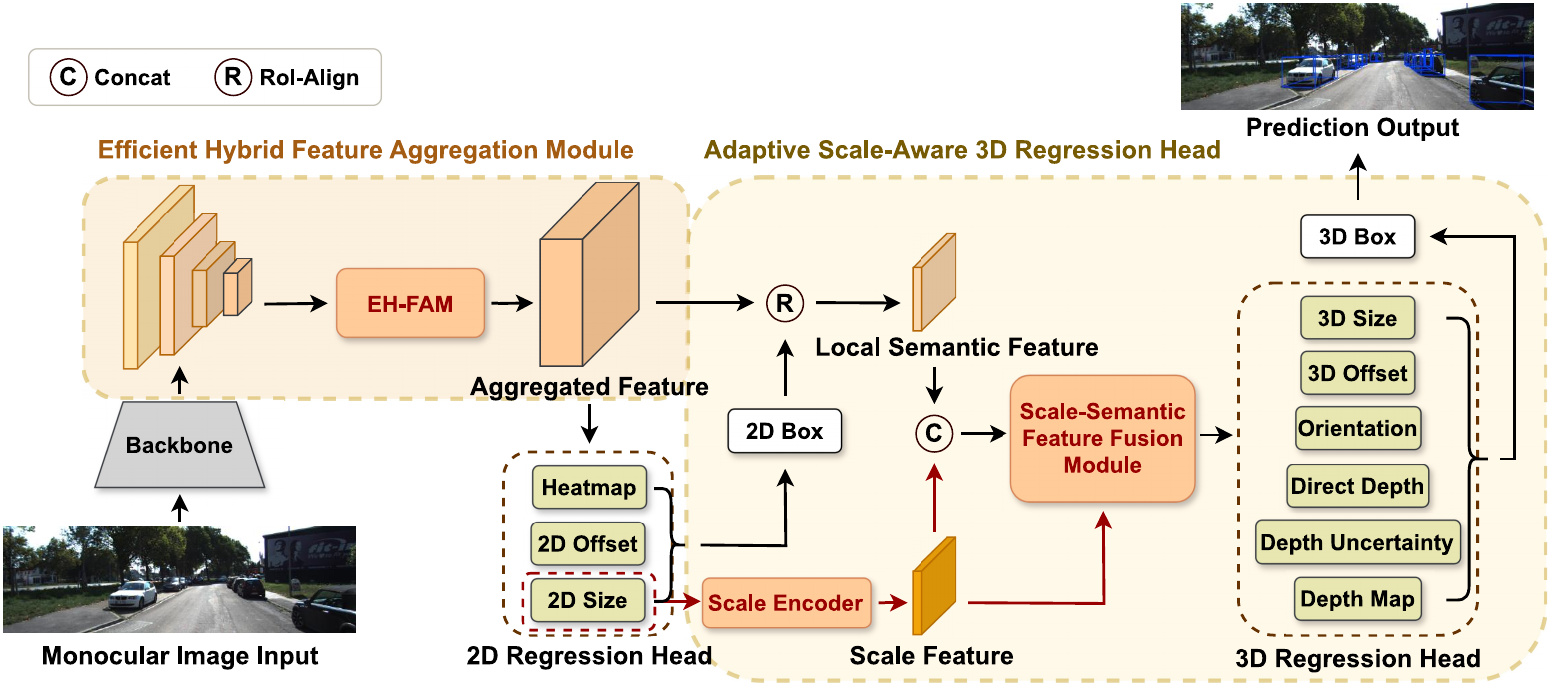}%
    \caption{
    Overview of our framework. The Efficient Hybrid Feature Aggregation Module (EH-FAM) efficiently aggregates multi-scale features. The Adaptive Scale-Aware 3D Regression Head (ASRH) fuses scale features with local semantics to guide the learning of 3D regression head.}
    \label{fig:overall}
    \vspace{-0.25cm}
\end{figure*}

In this section, we introduce our MonoASRH architecture. The overall framework, shown in Fig. \ref{fig:overall}, mainly consists of backbone, Efficient Hybrid Feature Aggregation Module, 2D regression head, and Adaptive Scale-Aware 3D Regression Head. Our pipeline mainly builds on the GUPNet \cite{lu2021geometry}. Detailed implementation will be discussed later in this section.

\subsection{Overall Architecture}
For a given RGB image $I \in {\mathbb{R}^{H \times W \times 3}}$, we extract multi-scale deep features using a pre-trained DLA-34 backbone\cite{yu2018deep}. These features, represented as $\left\{ {{S_1},{S_2},{S_3},{S_4}} \right\}$, undergo further refinement and cross-scale fusion through the proposed EH-FAM. The resulting aggregated features are denoted as $F \in {\mathbb{R}^{\frac{H}{4} \times \frac{W}{4} \times C}}$. Similar to\cite{lu2021geometry}, our 2D detector is based on CenterNet\cite{duan2019centernet}. 
We feed the deep features $F$ into three 2D detection heads to regress heatmaps $H \in {\mathbb{R}^{\frac{H}{4} \times \frac{W}{4} \times 3}}$, 2D offsets ${O_{2D}} \in {\mathbb{R}^{\frac{H}{4} \times \frac{W}{4} \times 2}}$, and 2D sizes ${S_{2D}} \in {\mathbb{R}^{\frac{H}{4} \times \frac{W}{4} \times 2}}$, enabling the prediction of 2D object centers and dimensions.

The deep features $F$ are also fed into ASRH for 3D attributes prediction. 
Specifically, RoI-Align extracts local semantic features ${F_{R}} \in {\mathbb{R}^{n \times d \times d \times C}}$ from $F$ based on 2D boxes, where $d \times d$ is the RoI-Align size and $n$ is the number of regions of interest.
Subsequently, we denote the size of 2D bounding box corresponding to each region of interest as ${S_{R}} \in {\mathbb{R}^{n \times 2}}$, which is further encoded into scale features.
Scale features and semantic features are then fused through the scale-semantic feature fusion module.
Ultimately, ASRH outputs 3D bounding box sizes ${S_{3D}} \in {\mathbb{R}^{n \times 3}}$, 3D center offsets ${O_{3D}} \in {\mathbb{R}^{n \times 2}}$, yaw angle $\Theta  \in {\mathbb{R}^{n \times 24}}$, direct depth $D \in {\mathbb{R}^{n \times d \times d}}$, and depth uncertainty ${D_U} \in {\mathbb{R}^{n \times d \times d}}$. Additionally, a depth attention map ${D_M} \in {\mathbb{R}^{n \times d \times d}}$ is employed to reduce the impact of irrelevant information, improving 3D detection accuracy.

\subsection{Efficient Hybrid Feature Aggregation Module}
\begin{figure}
    \centering
    \includegraphics[width=1.0\linewidth]{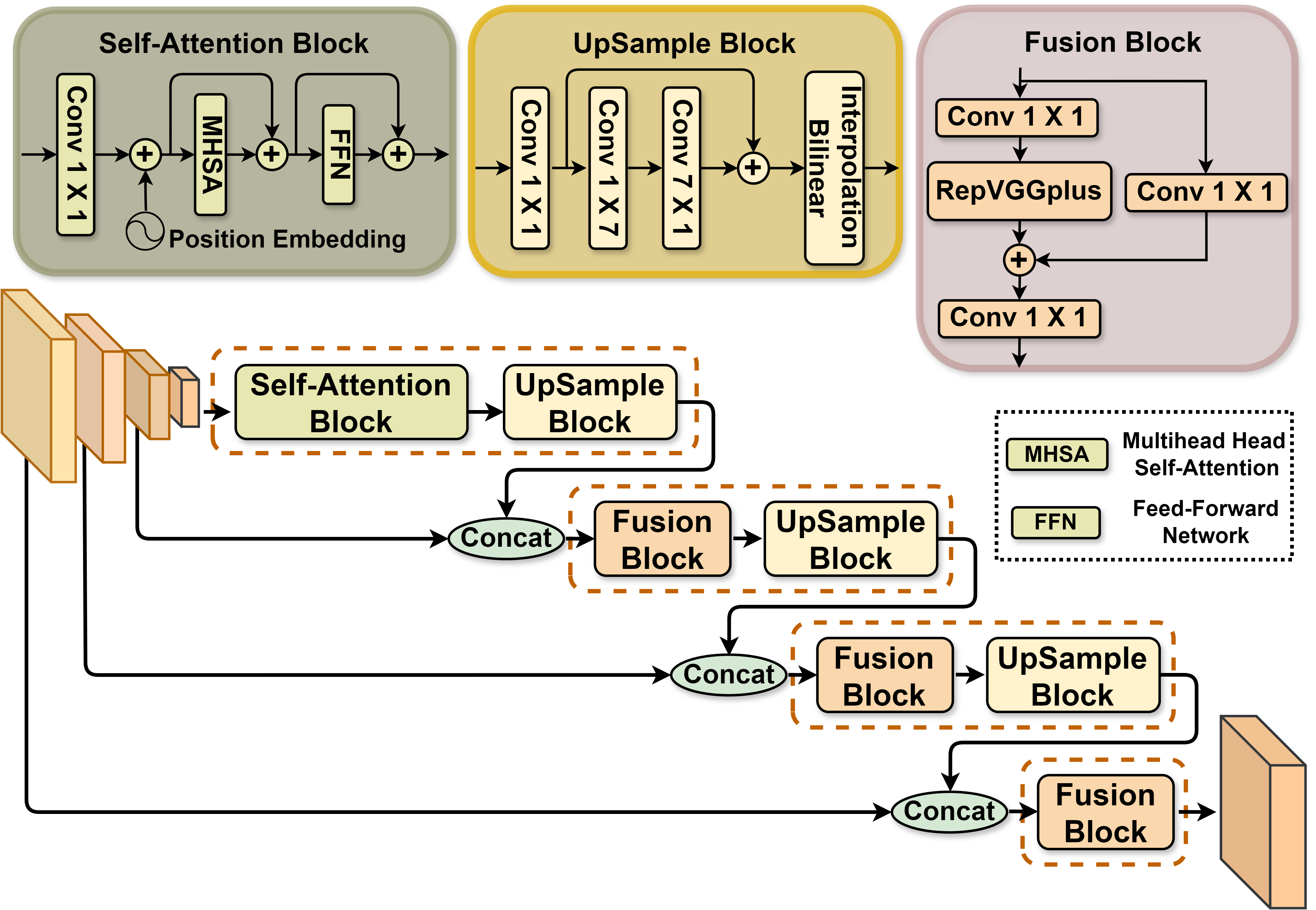}%
    \caption{Efficient Hybrid Feature Aggregation Module. Self-attention is applied only to the highest-level semantic feature map using an 8-head multi-head attention mechanism. Besides, all convolutional layers use the Mish activation function for improved feature representation.}
    \label{fig:EH-FAM}
    \vspace{-0.35cm}
\end{figure}

To efficiently aggregate features extracted by backbone at different scales into deep representations, we propose EH-FAM, a plug-and-play cross-scale feature aggregator. The ``hybrid'' module combines attention-based intra-scale interactions with CNN-based cross-scale fusion for improved performance and computational efficiency. The detailed implementation of EH-FAM is shown in Fig. \ref{fig:EH-FAM}.

\emph{1) Self-Attention Block:} The EH-FAM receives four feature maps with different downsampling ratios as input. 
To ensure global extraction of rich semantic information about objects, we apply multi-head self-attention to the highest-level feature ${S_4}$. We use separate linear projections for queries, keys, and values:
\begin{equation}
    \begin{array}{*{20}{c}}
    {{Q_i} = \left( {{X_{seq}} + {P_e}} \right)W_i^{\left( Q \right)},}&{{K_i} = \left( {{X_{seq}} + {P_e}} \right)W_i^{\left( K \right)},} 
    \end{array}
    \label{qk compute}
\end{equation}
\begin{equation}
    {V_i} = {X_{seq}}W_i^{\left( V \right)},
    \label{v compute}
\end{equation}
where ${{{X_{seq}}}} \in {\mathbb{R}^{T \times C}}$ obtained by flattening ${S_4}$, $T = \frac{H}{{32}} \times \frac{W}{{32}}$ represents sequence length, ${{P_e}}$ is the positional embedding vector, and $W_i^{\left( Q \right)},W_i^{\left( K \right)},W_i^{\left( V \right)}$ are the query, key, and value transformation matrices for the
$i$-th head, respectively.
For each head, the attention output is a weighted sum of the values:
\begin{equation}
    {Z_i} = \operatorname{Softmax} \left( {\frac{{{Q_i}K_i^T}}{{\sqrt {d_v} }}} \right){V_i}.
    \label{each head output}
\end{equation}
Finally, the concatenated output is passed through a linear layer to combine the multiple heads into a single output:
\begin{equation}
    {Z_{out}} = [{Z_1},{Z_2}, \ldots ,{Z_h}]{W_o},
    \label{multi-head attention output}
\end{equation}
where ${W_o} \in {\mathbb{R}^{\left( {h \times {d_v}} \right) \times C}}$ is a learnable projection matrix. $h$ and $d_v$ denote the number of attention heads and the channel dimensions of ${{V_i}}$, respectively. Next, the output of the multi-head attention is passed through an FFN and then reshaped back to its original spatial dimensions, denoted as ${S_{out}}$.

\emph{2) Upsampling Block:} To improve EH-FAM's ability to perceive small-scale objects, we design an upsample block and fusion block composed of lightweight convolutional modules. These blocks enable cross-scale information interaction and enhance the extraction of local semantic features. 
In the upsample block, we replace DLAUp's\cite{yu2018deep} large-kernel transposed convolutions with bilinear upsampling. 
To further mitigate information loss, we apply a 7$ \times $7 convolution prior to upsampling, which effectively refines feature maps. 
For efficiency, we decouple the 7$ \times $7 convolution into successive 1$ \times $7 and 7$ \times $1 convolutions, following the approach in\cite{szegedy2017inception}. 
Additionally, since some categories, such as Pedestrian and Cyclist, have relatively large aspect ratios but small sizes, decoupled convolutions in the horizontal and vertical directions are adopted to capture these elongated scale features. 

\begin{figure}
    \centering
    \includegraphics[width=0.8\linewidth]{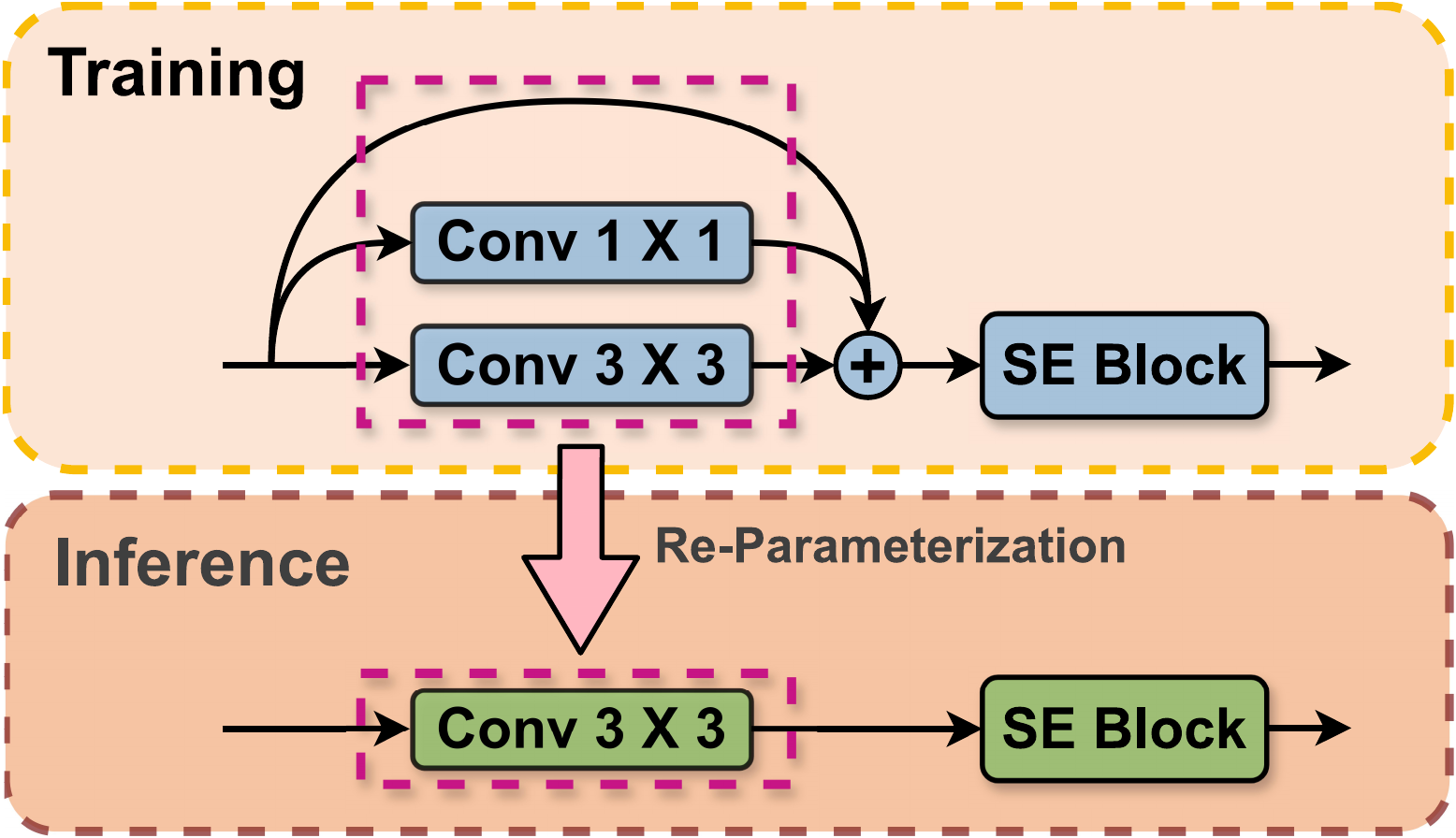}%
    \caption{Structural diagram of the RepVGGplus block. During training, RepVGGplus employs a multi-branch convolutional architecture, which is re-parameterized into a single 3$ \times $3 convolutional layer for inference.}
    \label{fig:repvgg}
    \vspace{-0.35cm}
\end{figure}

\emph{3) Fusion Block:} 
In the fusion block, we employ the RepVGGplus block\cite{ding2021repvgg}, which leverages re-parameterization techniques to convert a multi-branch ResNet-style architecture into a single-path VGG-like model during inference. The structure of the RepVGGplus block is illustrated in Fig. \ref{fig:repvgg}
This maintains the representational power of the model while accelerating inference. The final deep features are aggregated as follows:
\begin{equation}
    F = {f}\left( {{S_{out}},{S_3},{S_2},{S_1}} \right),
    \label{aggregated feature}
\end{equation}
where ${f}$ denotes the upsampling and cross-scale fusion operations.

\subsection{Adaptive Scale-Aware 3D Regression Head}
\begin{figure*}
    \centering
    \includegraphics[width=0.96\linewidth]{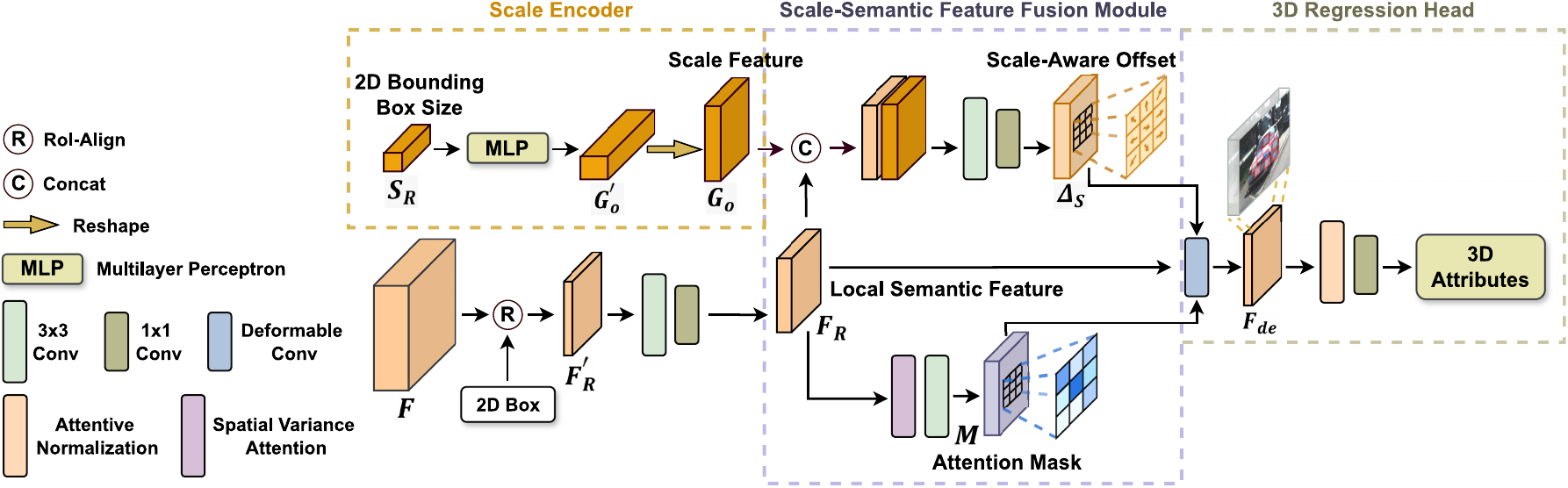}%
    \caption{The proposed Adaptive Scale-Aware 3D Regression Head. In Scale-Aware Offset, brown arrows indicate the direction of dynamic adjustment of the convolutional receptive field. In Attention Mask, varying shades of blue represent different weights at each position in the convolution kernel. 
    Our network includes six parallel 3D regression heads of this structure to regress different 3D attributes.}
    \label{fig:ASRH}
    \vspace{-0.35cm}
\end{figure*}

As illustrated in Fig. \ref{fig:ASRH}, the proposed Adaptive Scale-Aware 3D Regression Head decomposes the 3D bounding box regression process into three stages. First, scale features are captured by encoding the 2D bounding box dimensions. Next, a scale-semantic fusion module combines these scale features with semantic features extracted from the region of interest, dynamically adjusting the 3D regression head's receptive field. Given the limited image space occupied by foreground objects (e.g., the car class in KITTI covers just 11.42\% of depth pixels), we also introduce an attention mask to ensure the 3D regression head focuses on relevant foreground regions. Finally, the 3D regression head outputs the 3D bounding box attributes. 

\emph{1) Scale Encoder:} The Scale Encoder uses MLP to transform 2D geometric properties into a high-dimensional representation. By learning this transformation, the Scale Encoder effectively encodes scale information. For each region of interest, the corresponding 2D bounding box size ${S_{R}} \in {\mathbb{R}^{n \times 2}}$ is mapped to a high-dimensional feature space:
\begin{equation}
    {G'_o} = \operatorname{MLP} \left( {{S_R}} \right),
    \label{MLP for offset generated}
\end{equation}
where ${G'_o} \in {\mathbb{R}^{n \times 32}}$. 
Simultaneously, we apply a series of convolutional layers to further refine the feature maps ${F'_R}$ within the region of interest, obtaining the local semantic features ${F_R}$:
\begin{equation}
    {F_R} = \operatorname{Conv} \left( {\operatorname{ReLU} \left( {\operatorname{Conv} \left( {{F'_R}} \right)} \right)} \right),
    \label{local semantic features generated}
\end{equation}
where ${F_{R}} \in {\mathbb{R}^{n \times d \times d \times C}}$, $d \times d$ is the RoI-Align size. Offset $G'_o$ is then reshaped to match the spatial dimensions of the ${F_R}$, denoted as ${G_o}$.

\emph{2) Scale-Semantic Feature Fusion Module:} This module comprises two key components: offset feature generation and attention mask generation. For offset generation, the scale feature ${G_o}$ and local semantic features ${F_R}$ are concatenated. The combined features are processed through a stacked convolutional layer consisting of 3$ \times $3 and 1$ \times $1 convolutions, producing a scale-aware offset ${\Delta _S}$, which is used to dynamically adjust the receptive field in subsequent deformable convolutions. 
\begin{equation}
    {\Delta _S} = \operatorname{Conv} \left( {\operatorname{ReLU} \left( {\operatorname{Conv} \left( {\left[ {{G_o},{F_R}} \right]} \right)} \right)} \right),
    \label{scale-aware offset generation}
\end{equation}
where ${\Delta _S}  \in {\mathbb{R}^{n \times d \times d \times 18}}$, and 18 represents the offset of each spatial position in the deformable convolution kernel.

In the attention mask generation component, inspired by\cite{yang2021simam}, we design an attention mechanism based on spatial variance. This mechanism emphasizes regions with significant deviations from the mean, which are often foreground regions of interests.
First, the mean of the local semantic features ${F_{R}}$ are computed across spatial dimensions:
\begin{equation}
    \mu  = \frac{1}{{d \times d}}\sum\limits_{i = 1}^d {\sum\limits_{j = 1}^d {F_R^{n,i,j,c}} }.
    \label{mean for FR}
\end{equation}
Then, we calculate the squared difference of each pixel relative to the mean across the spatial dimensions of the feature map:
\begin{equation}
    V = {\left( {F_R^{n,i,j,c} - \mu } \right)^2}.
    \label{variance for each pixel}
\end{equation}
Using the squared difference obtained from the Eqn. \ref{variance for each pixel}, the attention weights can be calculated as:
\begin{equation}
    Y = \sigma \left( {\frac{V}{{{\frac{1}{{d \times d - 1}}\sum\limits_{i = 1}^d {\sum\limits_{j = 1}^d {{V_{i,j}} + \lambda } } } }} + 0.5} \right),
    \label{Attention Weights for FR}
\end{equation}
where $\lambda$ is a small positive constant. $\sigma$ is a projection function implemented by convolution followed by sigmoid. Normalizing by the variance provides a measure of how relatively important a specific deviation is compared to the overall distribution of deviations.

Finally, a convolutional layer with a sigmoid activation function is applied to generate the mask:
\begin{equation}
    M = \sigma \left( {{F_R} \odot Y} \right),
    \label{mask generate}
\end{equation}
where $M \in {\mathbb{R}^{n \times d \times d \times 9}}$, and $\odot$ denotes element-wise multiplication.

\emph{3) 3D Regression Head:} To achieve scale-aware dynamic receptive field adjustment, the first layer of the 3D regression head utilizes deformable convolutions\cite{zhu2019deformable}. As shown in Eqn. \ref{scale-aware deformable conv}, the offset ${\Delta _S}$ and attention mask $M$ generated by the Scale-Semantic Feature Fusion Module are applied to this layer, allowing the model to better detect and localize objects of varying scales:
\begin{equation}
    {F_{de}} = \sum\limits_{{P_n} \in \Omega } {{M_{{P_n}}} \cdot {W_{{P_n}}} \cdot {F_R}\left( {\Delta _S^{{P_n}} + {P_n}} \right)},
    \label{scale-aware deformable conv}
\end{equation}
where $\Omega$ represents the local neighborhood for a 3$ \times $3 kernel, ${{P_n}}$ enumerates the locations in $\Omega$, ${\Delta _S^{{P_n}}}$ is the learned offset for each position in the convolution kernel, ${{W_{{P_n}}}}$ is the weight of the convolution kernel at position ${{P_n}}$, and ${{M_{{P_n}}}}$ controls the contribution of the corresponding kernel position.

Next, we normalize the feature map ${F_{de}}$ using the Attentive Normalization layer\cite{li2020attentive}, which is a light weight module integrating feature normalization and channel-wise feature attention:
\begin{equation}
    {F_{n}} = \operatorname{AN} \left( {{F_{de}}} \right).
    \label{Attentive Normalization for Fde}
\end{equation}
Due to AN's mixture modeling approach to affine transformation, which recalibrates features after standardization, it is employed in the regression head to facilitate the learning of more expressive latent feature representations. Finally, the feature map is processed through a LeakyReLU activation function and a 1$ \times $1 convolution for channel mapping, ultimately regressing the 3D bounding box attributes.

\subsection{Loss Function}
\begin{figure}
    \centering
    \includegraphics[width=0.9\linewidth]{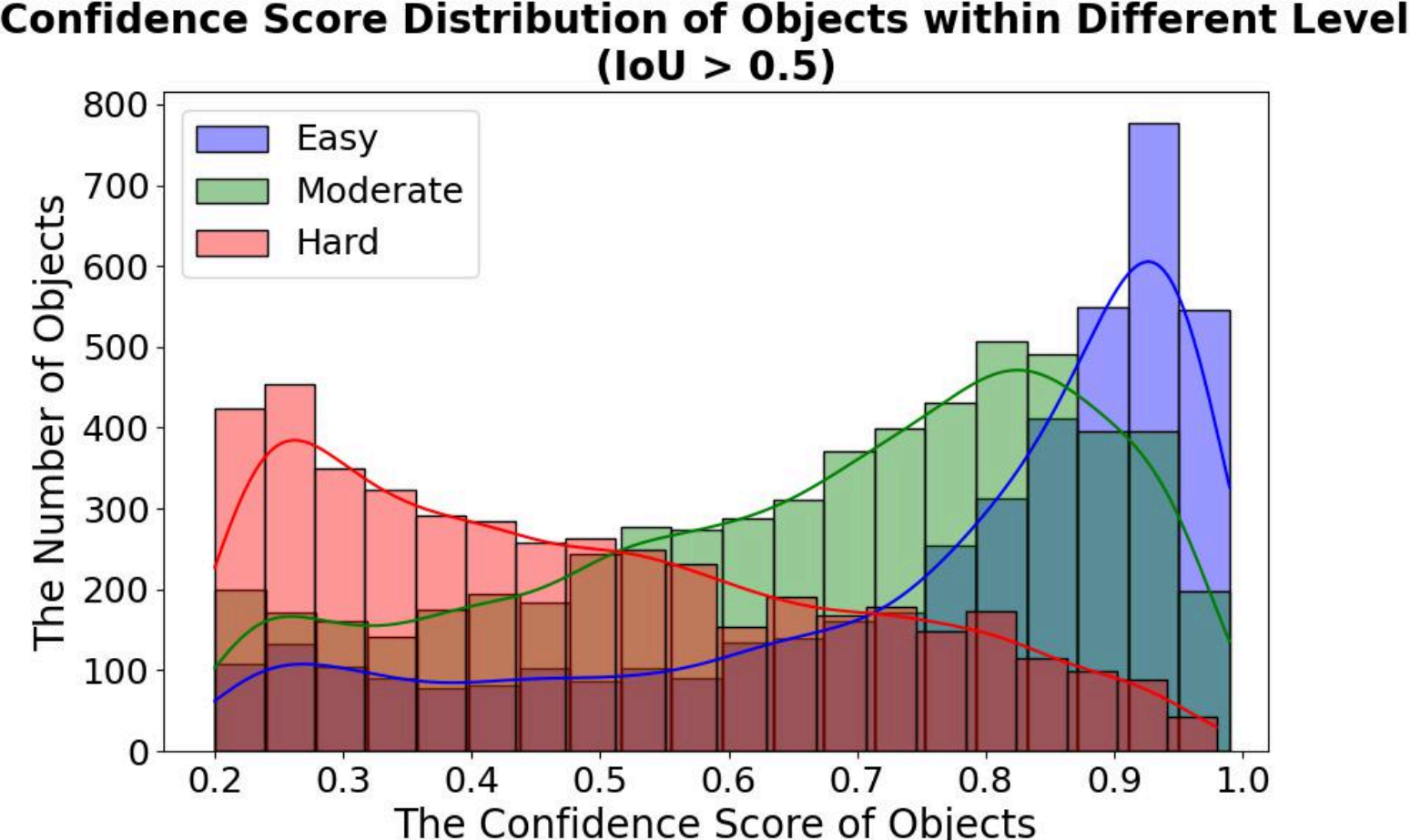}%
    \caption{We visualize the confidence score distribution of MonoASRH across three difficulty levels in the KITTI dataset, considering only samples with a 3D IoU greater than 0.5. The model tends to assign higher confidence scores to easy samples, while showing lower confidence for hard samples.}
    \label{fig:Hist}
    \vspace{-0.35cm}
\end{figure}

\emph{1) Selective Confidence-Guided Heatmap Loss:} 
Easy samples, which are typically fully visible and closer to the camera, exhibit less scale complexity. As a result, our scale-aware mechanism struggles to provide additional focus on these targets. To address this issue, we propose the Selective Confidence-Guided Heatmap Loss, which encourages the model to focus more on high-confidence samples, given that the network tends to assign higher confidence scores to easy samples (as shown in Fig. \ref{fig:Hist}).
Specifically, this loss function uses an improved version of Focal Loss\cite{law2018cornernet} as the primary loss 
and introduces a Selective Confidence-Guided (SCG) Loss as an auxiliary loss (Eqn. \ref{Selective Confidence-Guided Loss}) to enhance training.

\begin{equation}
    {\mathcal{L}_{SCG}} = - \frac{1}{B}\sum\limits_{i = 1}^B {\sum\limits_{j = 1}^K 
    \left\{
    \begin{aligned}
    &\log \left( {{t_{ij}}} \right),&{t_{ij}} > T \\
    &0,&{\text{otherwise}}\\
    \end{aligned}  
    \right. },
    \label{Selective Confidence-Guided Loss}
\end{equation}
where $B$ is the batch size, $K$ is the maximum number of detectable objects per image, ${t_{ij}}$ are the top $K$ points with the highest confidence in the predicted heatmap, and $T$ is the confidence threshold.

The focal loss introduces a dynamic weighting for heatmap regression.
The SCG Loss encourages the ASRH to prioritize high-confidence samples, which is critical for accurate detection of closer objects.
Meanwhile, optimizing for high-confidence objects can to some extent mitigate the overfitting of noise in low-confidence samples.
The final heatmap regression loss is a weighted combination of these two losses:
\begin{equation}
    {\mathcal{L}_{Heatmap}} = {\mathcal{L}_{focal}} + \lambda {\mathcal{L}_{SCG}}.
    \label{combined heatmap loss}
\end{equation}
In the experiment, 
$T$ was taken as 0.9 and $\lambda$ as 0.01.

\emph{2) Total Loss:} The overall network loss $\mathcal{L}$ consists of two components: 2D object loss ${\mathcal{L}_{2D}}$ and 3D object loss ${\mathcal{L}_{3D}}$. Mathematically, the composite loss function is expressed as:
\begin{equation}
    \mathcal{L} = {\mathcal{L}_{2D}} + {\mathcal{L}_{3D}}.
    \label{total loss}
\end{equation}

The 2D object loss follows the design of CenterNet\cite{duan2019centernet}. It includes the Selective Confidence-Guided Heatmap Loss ${\mathcal{L}_{Heatmap}}$, 2D center offset loss ${\mathcal{L}_{{O_{2D}}}}$ and 2D size loss ${\mathcal{L}_{{S_{2D}}}}$:
\begin{equation}
    {\mathcal{L}_{2D}} = {\mathcal{L}_{Heatmap}} + {\mathcal{L}_{{O_{2D}}}} + {\mathcal{L}_{{S_{2D}}}},
    \label{2D loss}
\end{equation}
where ${\mathcal{L}_{{O_{2D}}}}$ and ${\mathcal{L}_{{S_{2D}}}}$ use standard L1 loss.

Following the MonoDLE paradigm\cite{ma2021delving}, the 3D object loss trains the network to predict key 3D attributes. It includes the 3D center offset loss ${\mathcal{L}_{{O_{3D}}}}$ and 3D size loss ${\mathcal{L}_{{S_{3D}}}}$ with L1 loss, the orientation loss ${\mathcal{L}_\theta }$ using multi-bin loss\cite{mousavian20173d}, and the depth loss ${\mathcal{L}_{D}}$ supervised by Laplacian aleatoric uncertainty loss\cite{chen2020monopair}:
\begin{equation}
    {\mathcal{L}_{3D}} = {\mathcal{L}_{{O_{3D}}}} + {\mathcal{L}_{{S_{3D}}}} + {\mathcal{L}_\theta } + {\mathcal{L}_D}.
    \label{3D loss}
\end{equation}

\section{Experiments}
\subsection{Setup}
\emph{1) Dataset:} To evaluate the performance of our proposed method, we conducted experiments on the KITTI 3D object detection benchmark\cite{geiger2012we} and Waymo Open dataset\cite{sun2020scalability}. The KITTI dataset consists of 7,481 training images and 7,518 testing images. It includes three primary object classes: Car, Pedestrian, and Cyclist, with varying levels of difficulty (\emph{Easy}, \emph{Moderate}, and \emph{Hard}) based on factors such as object size, occlusion, and truncation. Following the previous studies\cite{chen20153d}, the training images are further split into a training set of 3,712 images and a validation set of 3,769 images.

The Waymo Open Dataset\cite{sun2020scalability} is a pivotal resource for autonomous driving scene understanding, featuring 798 training and 202 validation sequences, yielding approximately 160,000 and 40,000 samples, respectively. For our experiments, we adopt the approach from\cite{chen20153d}, generating 52,386 training images and 39,848 validation images by sampling every third frame from the training sequences. This dataset captures diverse real-world driving scenarios and categorizes objects into LEVEL\_1 and LEVEL\_2 based on LiDAR point density. 

\emph{2) Evaluation metrics:} In KITTI, model’s performance is typically measured using average precision in 3D space and birds-eye view ($A{P_{3D}}$ and $A{P_{BEV}}$) at 40 recall positions. The AP is calculated based on Intersection over Union between the predicted 3D bounding boxes and the ground truth, with specific IoU thresholds for each object class -- 0.7 for cars and 0.5 for pedestrians and cyclists. Following the official protocol\cite{geiger2012we}, we use $A{P_{3D}}$ and $A{P_{BEV}}$ on \emph{Moderate} category as main metrics. In Waymo, evaluations use two IoU thresholds (0.5 and 0.7) across four distance ranges: Overall, 0 - 30m, 30 - 50m, and 50m - $\infty $.

\emph{3) Data augmentation:} To enhance the generalization ability of our model, we incorporated several data augmentation techniques commonly used in object detection. During training, we applied random horizontal flipping and random cropping with a probability of 0.5, random scaling with a probability of 0.4, and random shifting with a probability of 0.1. Additionally, we utilized the MixUp3D technique from MonoLSS\cite{li2024monolss} as an extra data augmentation strategy to further improve the model's robustness against occlusions.

\emph{4) Implementation details:} Our proposed MonoASRH was trained on two RTX 3090 GPUs with batch size of 16. 
For the EH-FAM, we set the channel dimension ${d_v}$ of ${V_i}$ to 128 in the multi-head self-attention. Following\cite{lu2021geometry}, the RoI-Align size $d \times d$ is set to $7 \times 7$.
For KITTI, we trained the model for 450 epochs using the Adam optimizer\cite{kingma2014adam}, with a weight decay of $1e-5$. The initial learning rate was set to $1e-3$ and decayed by a factor of 10 at the 250th and 370th epochs.
For Waymo Open dataset, we trained the model for only 80 epochs with a batch size of 32, while keeping all other settings consistent with those used for KITTI.
During testing, the maximum number of detectable objects per image, $K$, was set to 50.

\subsection{Main Results}
We compared our proposed MonoASRH with state-of-the-art monocular 3D object detection methods on the KITTI test set for the car category. As shown in Table \ref{test on car category}, our method outperformed the CAD model-based DCD\cite{li2022densely}, even without additional data. Compared to other methods that do not use extra data, MonoASRH achieved improvements in $A{P_{3D|R40}}$ across all three difficulty levels by 0.65\%, 1.00\%, and 1.40\%. 
Experimental results validate the effectiveness of our method in detecting objects in the Car category.

\begin{table*}
\centering
\caption{Comparisons for the Car Category on the KITTI test set at $\operatorname{IoU}=0.7$. The Best and Second Best Results are Bolded and Underlined, Respectively. The Improvements are Shown in Purple Fonts. `\textsuperscript{†}' denote Reproduced Results.}
\footnotesize
\begin{center}
\tabcolsep=0.021\linewidth
\begin{tabular}{lccccccccc}
\toprule[0.15em]
\rule{0pt}{0.3cm}
\multirow{2}*{\centering \textbf{Methods}} & \multirow{2}*{\centering \textbf{Reference}} & \multirow{2}*{\centering \textbf{Extra Data}} & \multicolumn{3}{c}{\centering \textbf{\emph{Test,}} $A{P_{3D|R40}}$(\%)} & \multicolumn{3}{c}{\centering \textbf{\emph{Test,}} $A{P_{BEV|R40}}$(\%)} & \multirow{2}*{\makecell[c]{\textbf{Runtime}\\ \textbf{(ms)}}}\\ \cline{4-9}
\rule{0pt}{0.3cm}
& & & \textbf{Easy} & \textbf{Mod.} & \textbf{Hard} & \textbf{Easy} & \textbf{Mod.} & \textbf{Hard} \\
\midrule
\rule{0pt}{0.22cm}
D4LCN\cite{ding2020learning} & CVPR 2020 & Depth & 16.65 & 11.72 & 9.51 & 22.51 & 16.02 & 12.55 & -\\
\rule{0pt}{0.22cm}
Kinematic3D\cite{brazil2020kinematic} & ECCV 2020 & Multi-frames & 19.07 & 12.72 & 9.17 & 26.99 & 17.52 & 13.10 & 120\\
\rule{0pt}{0.22cm}
MonoRUn\cite{chen2021monorun} & CVPR 2021 & LiDAR & 19.65 & 12.30 & 10.58 & 27.94 & 17.34 & 15.24 & 70\\
\rule{0pt}{0.22cm}
CaDDN\cite{reading2021categorical} & CVPR 2021 & LiDAR & 19.17 & 13.41 & 11.46 & 27.94 & 18.91 & 17.19 & 630\\
\rule{0pt}{0.22cm}
AutoShape\cite{liu2021autoshape} & ICCV 2021 & CAD & 22.47 & 14.17 & 11.36 & 30.66 & 20.08 & 15.59 & -\\
\rule{0pt}{0.22cm}
MonoDTR\cite{huang2022monodtr} & CVPR 2022 & LiDAR & 21.99 & 15.39 & 12.73 & 28.59 & 20.38 & 17.14 & 37\\
\rule{0pt}{0.22cm}
DCD\cite{li2022densely} & ECCV 2022 & CAD & 23.81 & 15.90 & 13.21 & 32.55 & 21.50 & 18.25 & -\\
\midrule
\rule{0pt}{0.22cm}
SMOKE\cite{liu2020smoke} & CVPRW 2020 & None & 14.03 & 9.76 & 7.84 & 20.83 & 14.49 & 12.75 & 30\\
\rule{0pt}{0.2cm}
MonoDLE\cite{ma2021delving} & CVPR 2021 & None & 17.23 & 12.26 & 10.29 & 24.79 & 18.89 & 16.00 & 40\\
\rule{0pt}{0.22cm}
GUPNet\cite{lu2021geometry} & ICCV 2021 & None & 20.11 & 14.20 & 11.77 & - & - & - & 34\\
\rule{0pt}{0.22cm}
MonoGround\cite{qin2022monoground} & CVPR 2022 & None & 21.37 & 14.36 & 12.62 & 30.07 & 20.47 & 17.74 & 30\\
\rule{0pt}{0.22cm}
DEVIANT\cite{kumar2022deviant} & ECCV 2022 & None & 21.88 & 14.46 & 11.89 & 29.65 & 20.44 & 17.43 & 40\\
\rule{0pt}{0.22cm}
MonoCon\cite{liu2022learning} & AAAI 2022 & None & 22.50 & 16.46 & 13.95 & 31.12 & 22.10 & 19.00 & 26\\
\rule{0pt}{0.22cm}
MonoDDE\cite{li2022diversity} & CVPR 2022 & None & 24.93 & 17.14 & 15.10 & 33.58 & 23.46 & 20.37 & 40\\
\rule{0pt}{0.22cm}
PDR\cite{sheng2023pdr} & TCSVT 2023 & None & 23.69 & 16.14 & 13.78 & 31.76 & 21.74 & 18.79 & 29\\
\rule{0pt}{0.22cm}
DVDET\cite{hu2023aerial} & RAL 2023 & None & 23.19 & 15.44 & 13.07 & 32.05 & 22.15 & 19.32 & -\\
\rule{0pt}{0.22cm}
MonoDETR\cite{zhang2023monodetr} & ICCV 2023 & None & 25.00 & 16.47 & 13.58 & 33.60 & 22.11 & 18.60 & 38\\
\rule{0pt}{0.22cm}
MonoLSS\textsuperscript{†}\cite{li2024monolss} & 3DV 2024 & None & 24.91 & \underline{18.17} & \underline{15.52} & \textbf{34.90} & \underline{25.65} & \underline{22.45} & 35\\
\rule{0pt}{0.22cm}
MonoCD\cite{yan2024monocd} & CVPR 2024 & None & \underline{25.53} & 16.59 & 14.53 & 33.41 & 22.81 & 19.57 & 36\\
\rule{0pt}{0.22cm}
FD3D\cite{wu2024fd3d} & AAAI 2024 & None & 25.38 & 17.12 & 14.50 & \underline{34.20} & 23.72 & \underline{20.76} & 40\\
\midrule
\rowcolor{cyan!10}
\rule{0pt}{0.22cm}
\textbf{MonoASRH(Ours)} & - & None & \textbf{26.18} & \textbf{19.17} & \textbf{16.92} & \underline{34.84} & \textbf{25.89} & \textbf{22.48} & 32\\
\rule{0pt}{0.22cm}
\emph{Improvement} & - & \emph{v.s. second-best} & \textcolor[RGB]{208,77,145}{\textbf{+0.65}} & \textcolor[RGB]{208,77,145}{\textbf{+1.00}} & \textcolor[RGB]{208,77,145}{\textbf{+1.40}} & \textcolor[RGB]{208,77,145}{\textbf{-0.06}} & \textcolor[RGB]{208,77,145}{\textbf{+0.24}} & \textcolor[RGB]{208,77,145}{\textbf{+0.03}} & -\\
\bottomrule[0.15em]
\end{tabular}
\end{center}
\vspace{-0.5cm}
\label{test on car category}
\end{table*}

\begin{table}
\centering
\caption{Comparisons for the Pedestrian and Cyclist Category on the KITTI test set at $\operatorname{IoU}=0.5$. The Best and Second Best Results are Bolded and Underlined, Respectively.}
\scriptsize
\begin{center}
\tabcolsep=0.02\linewidth
\begin{tabular}{lccccccc}
\toprule[0.15em]
\rule{0pt}{0.25cm}
\multirow{3}*{\centering \textbf{Methods}} & \multirow{3}*{\makecell[c]{\textbf{Extra}\\ \textbf{Data}}} &
\multicolumn{6}{c}{\centering \textbf{\emph{Test,}} $A{P_{3D|R40}}$(\%)} \\
\cline{3-8}
\rule{0pt}{0.25cm}
& & \multicolumn{3}{c}{\centering \textbf{Pedestrian}} & \multicolumn{3}{c}{\centering \textbf{Cyclist}} \\
\cline{3-8}
\rule{0pt}{0.25cm}
& & {\centering \textbf{Easy}} & {\centering \textbf{Mod.}} & {\centering \textbf{Hard}} & {\centering \textbf{Easy}} & {\centering \textbf{Mod.}} & {\centering \textbf{Hard}} \\
\midrule
\rule{0pt}{0.21cm}
CaDDN\cite{reading2021categorical} & LiDAR & 12.87 & 8.14 & 6.76 & \underline{7.00} & 3.41 & 3.30 \\
\rule{0pt}{0.21cm}
MonoDTR\cite{huang2022monodtr} & LiDAR & \underline{15.33} & \underline{10.18} & \underline{8.61} & 5.05 & 3.27 & 3.19 \\
\rule{0pt}{0.21cm}
DCD\cite{li2022densely} & CAD & 10.37 & 6.73 & 6.28 & 4.27 & 2.74 & 2.41 \\
\midrule
\rule{0pt}{0.21cm}
DEVIANT\cite{kumar2022deviant} & None & 13.43 & 8.65 & 7.69 & 5.05 & 3.13 & 2.59 \\
\rule{0pt}{0.21cm}
MonoCon\cite{liu2022learning} & None & 13.10 & 8.41 & 6.94 & 2.80 & 1.92 & 1.55 \\
\rule{0pt}{0.21cm}
MonoDDE\cite{li2022diversity} & None & 11.13 & 7.32 & 6.67 & 5.94 & 3.78 & \underline{3.33} \\
\rule{0pt}{0.21cm}
PDR\cite{sheng2023pdr} & None & 11.61 & 7.72 & 6.40 & 2.72 & 1.57 & 1.50 \\
\rule{0pt}{0.21cm}
MonoDETR\cite{zhang2023monodetr} & None & 12.54 & 7.89 & 6.65 & \textbf{7.33} & \textbf{4.18} & 2.92 \\
\midrule
\rowcolor{cyan!10}
\rule{0pt}{0.21cm}
\textbf{MonoASRH(Ours)} & None & \textbf{16.90} & \textbf{11.24} & \textbf{9.64} & 6.49 & \underline{3.94} & \textbf{3.51} \\
\bottomrule[0.15em]
\end{tabular}
\end{center}
\vspace{-0.25cm}
\label{test on pedestrian and cyclist category}
\end{table}

\begin{table}
\centering
\caption{Performance of the Car Category on the KITTI validation set at $\operatorname{IoU}=0.7$. The Best and Second Best Results are Bolded and Underlined, Respectively.} 
\scriptsize
\begin{center}
\tabcolsep=0.025\linewidth
\begin{tabular}{lcccccc}
\toprule[0.15em]
\rule{0pt}{0.25cm}
\multirow{2}*{\centering \textbf{Methods}} & \multicolumn{3}{c}{\centering \textbf{\emph{Val,}} $A{P_{3D|R40}}$(\%)} & \multicolumn{3}{c}{\centering \textbf{\emph{Val,}} $A{P_{BEV|R40}}$(\%)} \\ 
\cline{2-7}
\rule{0pt}{0.25cm}
 & \textbf{Easy} & \textbf{Mod.} & \textbf{Hard} & \textbf{Easy} & \textbf{Mod.} & \textbf{Hard} \\
\midrule
\rule{0pt}{0.21cm}
GUPNet\cite{lu2021geometry} & 22.76 & 16.46 & 13.72 & 31.07 & 22.94 & 19.75 \\
\rule{0pt}{0.21cm}
DEVIANT\cite{kumar2022deviant} & 24.63 & 16.54 & 14.52 & 32.60 & 23.04 & 19.99 \\
\rule{0pt}{0.21cm}
MonoDDE\cite{li2022diversity} & 26.66 & 19.75 & 16.72 & 35.51 & 26.48 & 23.07 \\
\rule{0pt}{0.21cm}
PDR\cite{sheng2023pdr} & 27.65 & 19.44 & 16.24 & 35.59 & 25.72 & 21.35 \\
\rule{0pt}{0.21cm}
MonoDETR\cite{zhang2023monodetr} & \textbf{28.84} & \underline{20.61} & 16.38 & - & - & - \\
\rule{0pt}{0.21cm}
MonoCD\cite{yan2024monocd} & 26.45 & 19.37 & 16.38 & 34.60 & 24.96 & 21.51 \\
\rule{0pt}{0.21cm}
MonoLSS\cite{li2024monolss} & 25.91 & 18.29 & 15.94 & 34.70 & 25.36 & 21.84 \\
\rule{0pt}{0.21cm}
FD3D\cite{wu2024fd3d} & 28.22 & 20.23 & \underline{17.04} & \underline{36.98} & \underline{26.77} & \underline{23.16} \\
\midrule
\rowcolor{cyan!10}
\rule{0pt}{0.21cm}
\textbf{MonoASRH(Ours)} & \underline{28.35} & \textbf{20.75} & \textbf{17.56} & \textbf{37.53} & \textbf{27.26} & \textbf{24.16} \\
\bottomrule[0.15em]
\end{tabular}
\end{center}
\vspace{-0.25cm}
\label{val on car category}
\end{table}

The results for the Pedestrian and Cyclist categories in the KITTI test set are presented in Table \ref{test on pedestrian and cyclist category}. For Pedestrians, MonoASRH achieved the best performance across all difficulty levels, outperforming the LiDAR-based MonoDTR method\cite{huang2022monodtr}. In the Cyclist category, MonoASRH ranked third at the \emph{Easy} level but still showed competitive performance, with a slight drop at the \emph{Moderate} level compared to the Transformer-based MonoDETR\cite{zhang2023monodetr}. However, it achieved the best performance at the \emph{Hard} level with a 3.51\% $A{P_{3D|R40}}$. 
These results highlight the challenges of detecting smaller, non-rigid objects like Pedestrians and Cyclists. Nonetheless, MonoASRH's scale-aware dynamic receptive field adjustment capability enabled strong performance in these categories, validating the model's generalization and scalability. 
Additionally, the decoupled convolutional layers in EH-FAM are better equipped to capture the features of these categories with large aspect ratios.

\begin{figure*}
    \centering
    \includegraphics[width=1.0\linewidth]{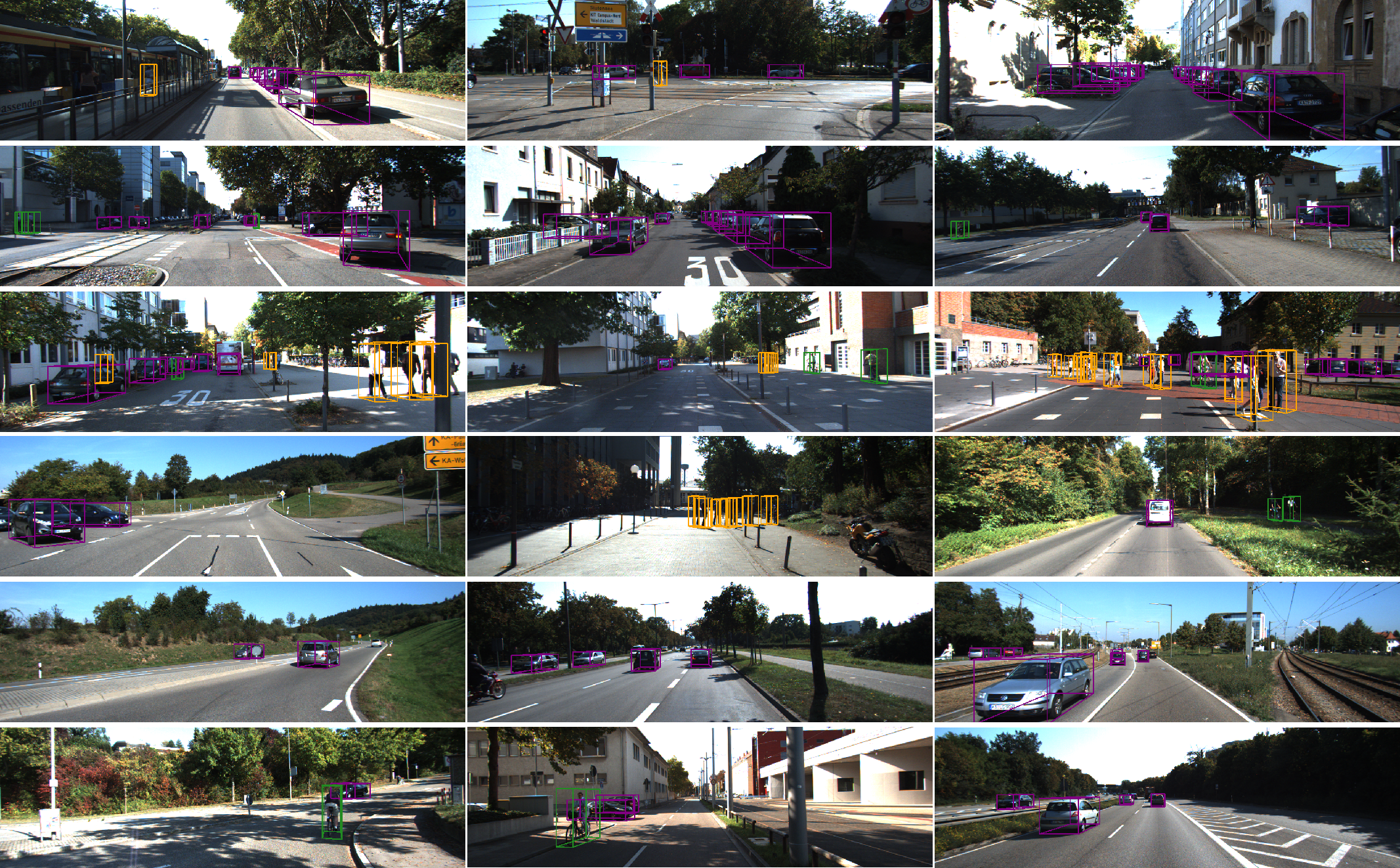}%
    \caption{Qualitative results produced by the proposed MonoASRH on the KITTI test set. The detected 3D bounding boxes for Car, Pedestrian and Cyclist are shown in purple, orange and green, respectively.}
    \label{fig:qualitative results testset}
    \vspace{-0.1cm}
\end{figure*}

\begin{table*}
\centering
\caption{Comparisons for the Vehicle Category on the Waymo Open validation set. We Use $A{P_{3D}}$ (LEVEL\_1 and LEVEL\_2, IoU $>$ 0.5 and IoU $>$ 0.7) according to Three Object Distance Intervals. `\textsuperscript{†}' denote Reproduced Results.}
\scriptsize
\begin{center}
\tabcolsep=0.0040\linewidth
\begin{tabular}{lccccccccccccccccc}
\toprule[0.15em]
\rule{0pt}{0.25cm}
\multirow{2}*{\centering \textbf{Methods}} & \multirow{2}*{\centering \textbf{Reference}} & \multicolumn{4}{c}{\centering \textbf{LEVEL\_1}(IoU $>$ 0.5)} & \multicolumn{4}{c}{\centering \textbf{LEVEL\_2}(IoU $>$ 0.5)} & \multicolumn{4}{c}{\centering \textbf{LEVEL\_1}(IoU $>$ 0.7)} & \multicolumn{4}{c}{\centering \textbf{LEVEL\_2}(IoU $>$ 0.7)} \\
\cline{3-18}
\rule{0pt}{0.25cm}
& & \textbf{Overall} & \textbf{0-30m} & \textbf{30-50m} & \textbf{50m-Inf} & \textbf{Overall} & \textbf{0-30m} & \textbf{30-50m} & \textbf{50m-Inf} & \textbf{Overall} & \textbf{0-30m} & \textbf{30-50m} & \textbf{50m-Inf} & \textbf{Overall} & \textbf{0-30m} & \textbf{30-50m} & \textbf{50m-Inf} \\
\midrule
\rule{0pt}{0.21cm}
With extra data: & & & & & & & & & & & & & & & & & \\
\rule{0pt}{0.21cm}
PatchNet\cite{ma2020rethinking} & ECCV 2020 & 2.92 & 10.03 & 1.09 & 0.23 & 2.42 & 10.01 & 1.07 & 0.22 & 0.39 & 1.67 & 0.13 & 0.03 & 0.38 & 1.67 & 0.13 & 0.03 \\
\rule{0pt}{0.21cm}
PCT\cite{wang2021progressive} & NIPS 2021 & 4.20 & 14.70 & 1.78 & 0.39 & 4.03 & 14.67 & 1.74 & 0.36 & 0.89 & 3.18 & 0.27 & 0.07 & 0.66 & 3.18 & 0.27 & 0.07 \\
\rule{0pt}{0.21cm}
NF-DVT\cite{pan2024depth} & JAS 2024 & 11.32 & 23.35 & \underline{5.76} & \underline{0.98} & \underline{11.24} & 23.20 & \underline{5.68} & \underline{0.92} & 2.76 & 5.89 & 0.99 & \underline{0.11} & 2.64 & 5.85 & \underline{0.96} & 0.10 \\
\midrule
\rule{0pt}{0.21cm}
Without extra data: & & & & & & & & & & & & & & & & & \\
\rule{0pt}{0.21cm}
GUPNet\cite{lu2021geometry} & ICCV 2021 & 10.02 & 24.78 & 4.84 & 0.22 & 9.39 & 24.69 & 4.67 & 0.19 & 2.28 & 6.15 & 0.81 & 0.03 & 2.14 & 6.13 & 0.78 & 0.02 \\
\rule{0pt}{0.21cm}
MonoJSG\cite{lian2022monojsg} & CVPR 2022 & 5.65 & 20.86 & 3.91 & 0.97 & 5.34 & 20.79 & 3.79 & 0.85 & 0.97 & 4.65 & 0.55 & 0.10 & 0.91 & 4.64 & 0.55 & 0.09 \\
\rule{0pt}{0.21cm}
DEVIANT\cite{kumar2022deviant} & ECCV 2022 & 10.98 & 26.85 & 5.13 & 0.18 & 10.29 & 26.75 & 4.95 & 0.16 & 2.69 & 6.95 & 0.99 & 0.02 & 2.52 & 6.93 & 0.95 & 0.02 \\
\rule{0pt}{0.21cm}
MonoRCNN++\cite{shi2023multivariate} & WACV 2023 & 11.37 & \underline{27.95} & 4.07 & 0.42 & 10.79 & \underline{27.88} & 3.98 & 0.39 & \textbf{4.28} & \textbf{9.84} & 0.91 & 0.09 & \textbf{4.05} & \textbf{9.81} & 0.89 & 0.08 \\
\rule{0pt}{0.21cm}
MonoUNI\cite{jinrang2023monouni} & NIPS 2023 & 10.98 & 26.63 & 4.04 & 0.57 & 10.38 & 26.57 & 3.95 & 0.53 & 3.20 & 8.61 & 0.87 & \textbf{0.13} & 3.04 & 8.59 & 0.85 & \textbf{0.12} \\
\rule{0pt}{0.21cm}
MonoCD\textsuperscript{†}\cite{yan2024monocd} & CVPR 2024 & \underline{11.62} & 27.21 & 5.24 & 0.83 & 11.14 & 27.14 & 5.12 & 0.75 & \underline{3.85} & 8.43 & \underline{1.00} & \textbf{0.13} & \underline{3.50} & 7.53 & 0.87 & \underline{0.11} \\
\rowcolor{cyan!10}
\rule{0pt}{0.21cm}
\textbf{MonoASRH(Ours)} & - & \textbf{12.35} & \textbf{29.25} & \textbf{6.16} & \textbf{1.10} & \textbf{11.58} & \textbf{29.14} & \textbf{5.95} & \textbf{0.96} & 3.36 & \underline{9.49} & \textbf{1.01} & \textbf{0.13} & 3.15 & \underline{9.46} & \textbf{0.97} & \underline{0.11} \\
\bottomrule[0.15em]
\end{tabular}
\end{center}
\vspace{-0.5cm}
\label{val on waymo vehicle category}
\end{table*}

As shown in Table \ref{val on car category}, MonoASRH was also evaluated on the KITTI validation set. For the Car category at the 0.7 IoU threshold and \emph{Moderate} difficulty, MonoASRH achieved state-of-the-art performance with $A{P_{3D|R40}}$ and $A{P_{BEV|R40}}$ scores of 20.75\% and 27.26\%, respectively. Some qualitative results on the KITTI test set are shown in Fig. \ref{fig:qualitative results testset}. Benefiting from the globally-aware self-attention mechanism in EH-FAM, the model effectively captures long-range dependencies for smaller objects such as distant cars. 

Table \ref{val on waymo vehicle category} compares the $AP_{3D}$ metrics for the ``Vehicle" category across various methods on the Waymo Open Val Set. 
In the 30-50m range, our model achieved state-of-the-art performance. Compared to NF-DVT\cite{pan2024depth}, MonoASRH improved the $AP_{3D}$ by 0.4\% and 0.02\% at IoU thresholds of 0.5 and 0.7, respectively, within the 30-50m range under the LEVEL\_1 setting. Additionally, under the LEVEL\_2 setting, the $AP_{3D}$ metrics improved by 0.27\% and 0.01\% at different IoU thresholds within the 30-50m range. These results further validate the effectiveness of MonoASRH in detecting distant and small objects.

\subsection{Ablation Studies}
\begin{figure}
    \centering
    \includegraphics[width=1.0\linewidth]{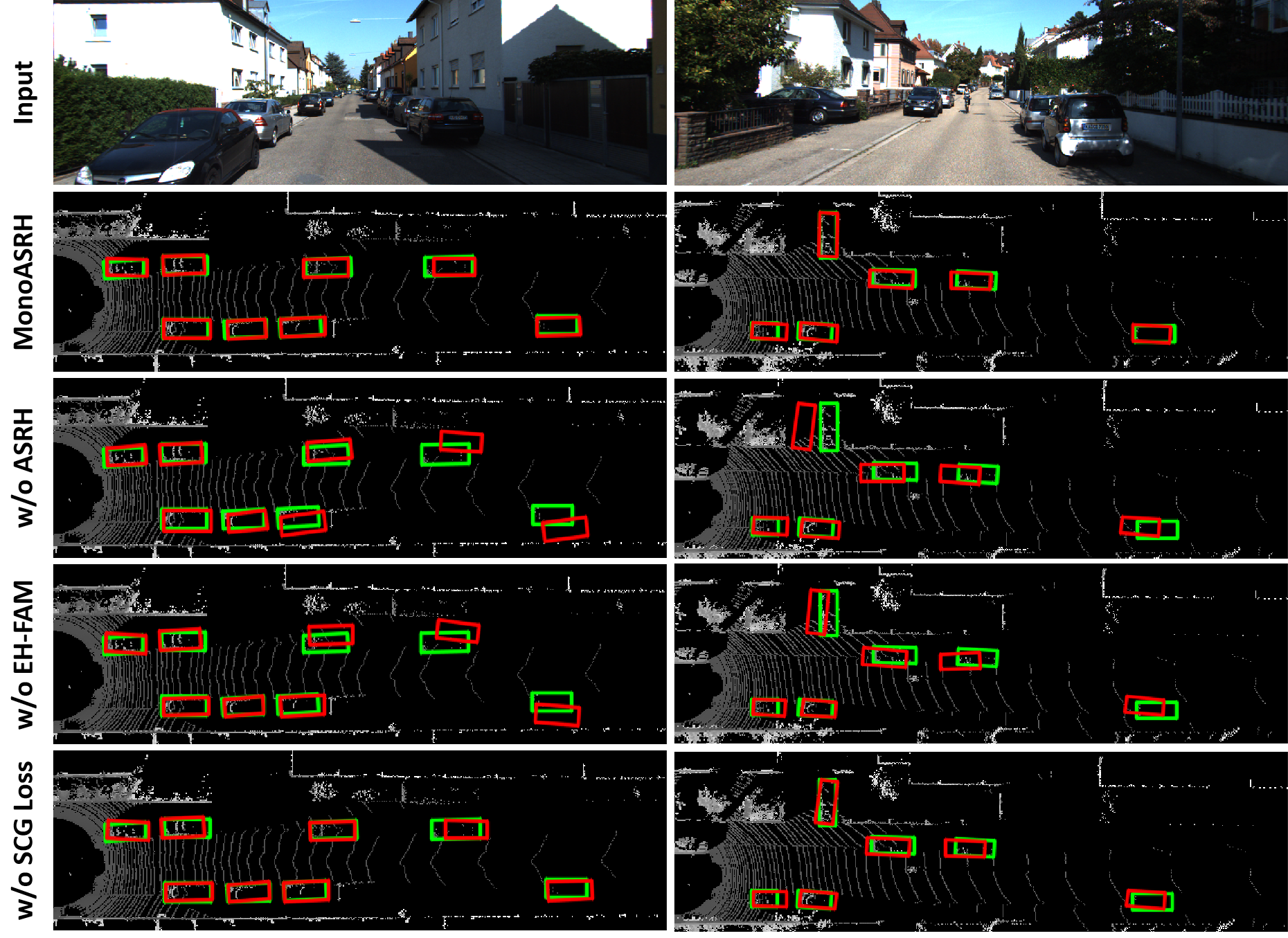}%
    \caption{Qualitative comparison results of ablation study on different components of our overall framework in the BEV view. Predicted and ground truth bounding boxes are shown in red and green, respectively.}
    \label{fig:Ablation1}
    \vspace{-0.2cm}
\end{figure}

In this section, we evaluate the impact of each component of the proposed framework on the network's performance. Ablation studies were conducted on the KITTI validation set using the Car category, with $A{P_{3D|R40}}$ and $A{P_{BEV|R40}}$ as metrics. Our baseline model is based on the GUPNet\cite{lu2021geometry} and incorporates the Learnable Sample Selection from MonoLSS\cite{li2024monolss} for initial training. We compare the effects of different components, including the Adaptive Scale-Aware 3D Regression Head, Efficient Hybrid Feature Aggregation Module, and Selective Confidence-Guided Heatmap Loss. 
As shown in Fig. \ref{fig:Ablation1}, we provide detailed visual comparisons of various approaches when the model is not equipped with a specific module.
The results in Table \ref{ablation total} and Fig. \ref{fig:Ablation1} demonstrate the contributions of each module to the overall performance of our method.

\begin{table}
\centering
\caption{Ablation Study on Different Components of our Overall Framework on KITTI val set for Car Category. `$A$', `$E$' and `${\mathcal{L}_{H}}$' denote Adaptive Scale-Aware 3D Regression Head, Efficient Hybrid Feature Aggregation Module, and Selective Confidence-Guided Heatmap Loss, Respectively.}
\begin{center}
\scriptsize
\tabcolsep=0.05\linewidth
\begin{tabular}{cccccc}
\toprule[0.15em]
\multicolumn{1}{c}{\rule{0pt}{0.25cm} \multirow{2}{*}{\centering $A$}} & \multirow{2}{*}{\centering $E$} & \multirow{2}{*}{\centering ${\mathcal{L}_{H}}$} 
& \multicolumn{3}{c}{\centering \textbf{\emph{Val,}} $A{P_{3D|R40}}$(\%)} \\ \cline{4-6} 
\rule{0pt}{0.25cm}
& & & \textbf{{Easy}} & \textbf{Mod.} & \textbf{{Hard}} \\ 
\midrule
\rule{0pt}{0.23cm}
- & - & - & 25.30 & 18.20 & 15.80 \\ 
\rule{0pt}{0.23cm}
\checkmark & - & - & 26.12 & 20.32 & 17.34 \\ 
\rule{0pt}{0.23cm}
- & \checkmark & - & 26.67 & 19.39 & 16.11 \\
\rule{0pt}{0.23cm}
- & - & \checkmark & 25.87 & 18.23 & 15.76 \\
\rule{0pt}{0.23cm}
\checkmark & \checkmark & - & 27.98 & 20.73 & 17.45 \\
\rowcolor{cyan!10}
\rule{0pt}{0.23cm}
\checkmark & \checkmark & \checkmark & \textbf{28.35} & \textbf{20.75} & \textbf{17.56} \\
\bottomrule[0.15em]
\end{tabular}
\end{center}
\vspace{-0.3cm}
\label{ablation total}
\end{table}

\begin{figure}
    \centering
    \includegraphics[width=1.0\linewidth]{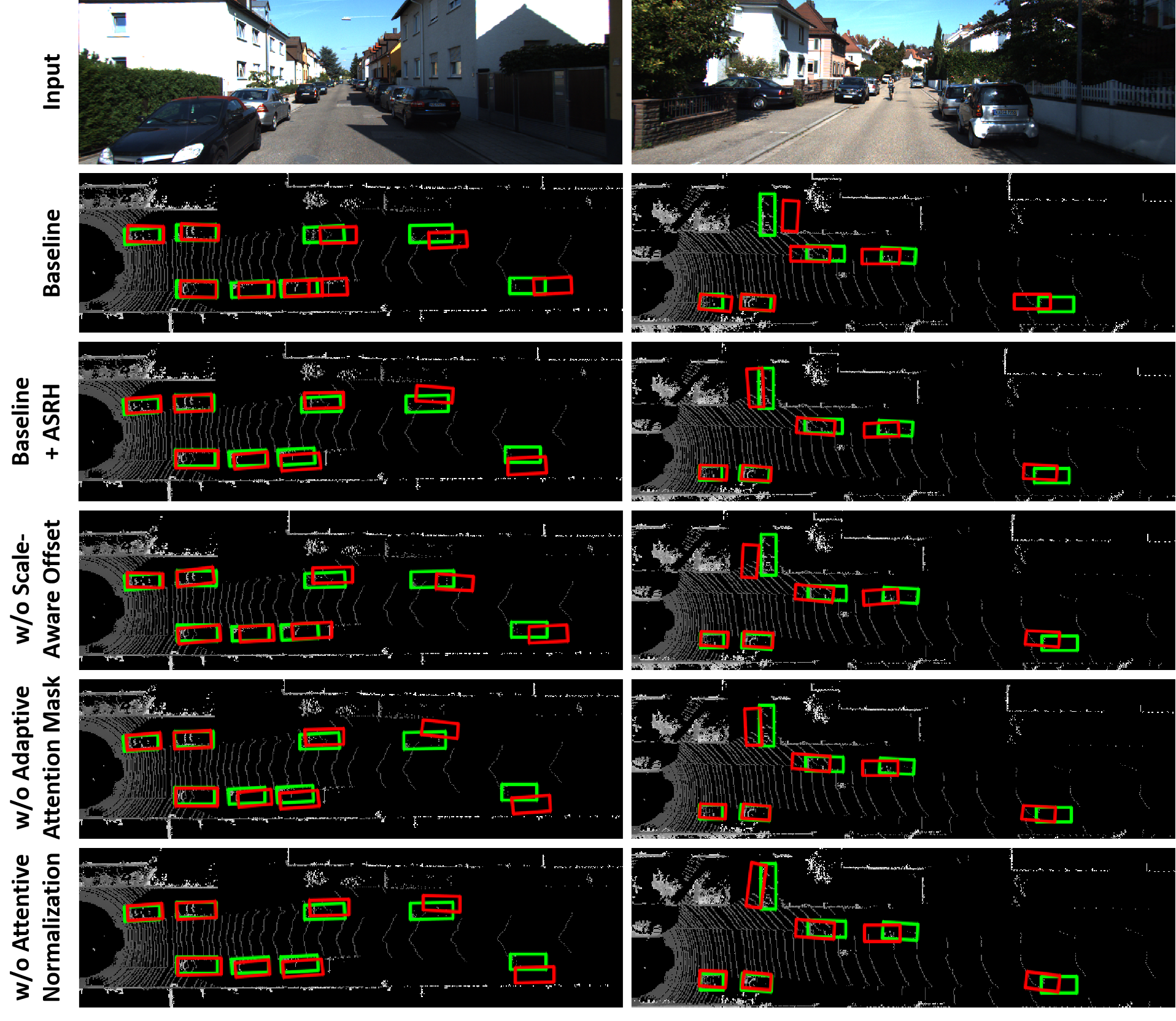}%
    \caption{Qualitative comparison results of ablation study on ASRH in the BEV view. Predicted and ground truth bounding boxes are shown in red and green, respectively.}
    \label{fig:Ablation2}
    \vspace{-0.2cm}
\end{figure}

\emph{1) Adaptive Scale-Aware 3D Regression Head:} In Table \ref{ablation ASRH}, we further explore the contribution of each component within the Adaptive Scale-Aware 3D Regression Head to model's performance. Specifically, we evaluated ASRH under three settings: ``w/o Scale-Aware Offset", where receptive field offsets are derived solely from semantic features without object scale information; ``w/o Adaptive Attention Mask", where the attention mask is randomly initialized; ``w/o Attentive Normalization", where the AN layer\cite{li2020attentive} is not applied to normalize feature maps ${F_{de}}$. The experimental results show that incorporating scale information is crucial in ASRH, enabling the model to perform dynamic scale-aware adjustments of the receptive field. This significantly improves detection, particularly for small and occluded objects at \emph{Moderate} and \emph{Hard} difficulty levels.
We also provide detailed visual comparisons in Fig. \ref{fig:Ablation2}, which further demonstrate the role of each component in ASRH.

\begin{table}
\centering
\caption{Ablation Study of ASRH on KITTI val set for Car Category.}
\begin{center}
\scriptsize
\tabcolsep=0.04\linewidth
\begin{tabular}{cccc}
\toprule[0.15em]
\rule{0pt}{0.25cm}
\multirow{2}*{\centering \textbf{Architecture}} & \multicolumn{3}{c}{\centering \textbf{\emph{Val,}} $A{P_{3D|R40}}$(\%)} \\ \cline{2-4}
\rule{0pt}{0.25cm}
& {\centering \textbf{Easy}} & {\centering \textbf{Mod.}} & {\centering \textbf{Hard}} \\
\midrule
\rowcolor{cyan!10}
\rule{0pt}{0.23cm}
\textbf{Baseline + ASRH} & \textbf{26.12} & \textbf{20.32} & \textbf{17.34} \\
\rule{0pt}{0.23cm}
w/o Scale-Aware Offset & 25.72 & 18.91 & 16.35 \\
\rule{0pt}{0.23cm}
w/o Adaptive Attention Mask & 25.63 & 19.86 & 17.00 \\
\rule{0pt}{0.23cm}
w/o Attentive Normalization & 25.97 & 20.15 & 17.16 \\
\bottomrule[0.15em]
\end{tabular}
\end{center}
\vspace{-0.3cm}
\label{ablation ASRH}
\end{table}

Additionally, we compared ASRH with the standard deformable convolution\cite{zhu2019deformable}. We inserted deformable convolution into the first layer of each 3D regression head in the baseline model and conducted comparison experiments with ASRH. As shown in Table \ref{Comparative Experiments ASRH}, ASRH consistently outperformed standard deformable convolutions in monocular 3D detection tasks.

\begin{table}
\centering
\caption{Comparison Experiment between ASRH and Deformable Convolution on KITTI val set for Car Category. `$\operatorname{Deform}.$' Denotes Deformable Convolution.}
\begin{center}
\scriptsize
\tabcolsep=0.025\linewidth
\begin{tabular}{lcccccc}
\toprule[0.15em]
\rule{0pt}{0.25cm}
\multirow{2}*{\textbf{Methods}} & \multicolumn{3}{c}{\centering \textbf{\emph{Val,}} $A{P_{3D|R40}}$(\%)} & \multicolumn{3}{c}{\centering \textbf{\emph{Val,}} $A{P_{BEV|R40}}$(\%)} \\ \cline{2-4} \cline{5-7} 
\rule{0pt}{0.25cm}
& {\centering \textbf{Easy}} & {\centering \textbf{Mod.}} & {\centering \textbf{Hard}} & {\centering \textbf{Easy}} & {\centering \textbf{Mod.}} & {\centering \textbf{Hard}} \\
\midrule
\rule{0pt}{0.23cm}
Baseline & 25.30 & 18.20 & 15.80 & 34.41 & 25.50 & 21.86 \\
\rule{0pt}{0.23cm}
+ Deform. & 25.71 & 18.85 & 16.88 & 34.55 & 25.80 & 22.27 \\
\rowcolor{cyan!10}
\rule{0pt}{0.23cm}
+ ASRH & \textbf{26.12} & \textbf{20.32} & \textbf{17.34} & \textbf{35.26} & \textbf{27.36} & \textbf{23.87} \\
\bottomrule[0.15em]
\end{tabular}
\end{center}
\vspace{-0.2cm}
\label{Comparative Experiments ASRH}
\end{table}

\emph{2) Efficient Hybrid Feature Aggregation Module:} 
In order to validate EH-FAM’s generalizability, we replaced the original feature aggregation layers in three baseline models (MonoDLE\cite{ma2021delving}, GUPNet\cite{lu2021geometry}, and MonoLSS\cite{li2024monolss}) with our EH-FAM:
\begin{itemize}
    \item MonoDLE uses a stacked hourglass network for multi-scale feature fusion. We replaced its final feature aggregation block with EH-FAM while retaining the backbone and detection heads.
    \item GUPNet employs DLAUp\cite{yu2018deep}, a hierarchical aggregation module. We substituted DLAUp entirely with EH-FAM, preserving the backbone and 3D regression head.
    \item MonoLSS also adopts the DLAUp structure. We replaced it with EH-FAM, keeping the Learnable Sample Selection module unchanged.
\end{itemize}

No additional components were added or removed beyond this substitution. The baseline models’ training protocols, loss functions, and hyperparameters remained identical to ensure fair comparison.

The results in Table \ref{ablation EH-FAM} show that EH-FAM led to performance improvements and parameter reduction across all three baselines. Specifically, the $A{P_{3D|R40}}$ scores for MonoDLE, GUPNet, and MonoLSS on the \emph{Moderate} level increased by 0.75\%, 0.05\%, and 0.91\%, respectively, while their parameter counts decreased by 8.76\%, 8.74\%, and 8.48\%. Additionally, the computational cost was reduced by 5.71\%, 7.25\%, and 7.17\%, respectively. These results confirm that EH-FAM can be widely applied to various baselines, significantly enhancing both performance and computational efficiency.

\begin{table}
\centering
\caption{Ablation Study of EH-FAM with Different Baselines on KITTI val set for Car Category. `$\oplus$' Indicates that the Feature Aggregation Module of the Original Baseline Model is Replaced with EH-FAM.}
\begin{center}
\scriptsize
\tabcolsep=0.02\linewidth
\begin{tabular}{lccccc}
\toprule[0.15em]
\rule{0pt}{0.25cm}
\multirow{2}*{\centering \textbf{Methods}} & \multicolumn{3}{c}{\centering \textbf{\emph{Val,}} $A{P_{3D|R40}}$(\%)} & \multirow{2}*{\centering \textbf{Params(M)}$\downarrow$} & \multirow{2}*{\centering \textbf{FLOPs(G)}$\downarrow$} \\ \cline{2-4}
\rule{0pt}{0.25cm}
& {\centering \textbf{Easy}} & {\centering \textbf{Mod.}} & {\centering \textbf{Hard}} & & \\
\midrule
\rule{0pt}{0.23cm}
MonoDLE\cite{ma2021delving} & 17.45 & 13.66 & 11.68 & 19.86 & 158.18 \\
\rowcolor{cyan!10}
\rule{0pt}{0.23cm}
MonoDLE $\oplus$ EH-FAM & \textbf{19.89} & \textbf{14.41} & \textbf{12.35} & \textbf{18.12} & \textbf{149.14} \\
\rule{0pt}{0.23cm}
\emph{Improvement} & \textcolor[RGB]{208,77,145}{\textbf{+2.44}} & \textcolor[RGB]{208,77,145}{\textbf{+0.75}} & \textcolor[RGB]{208,77,145}{\textbf{+0.67}} & \textcolor[RGB]{208,77,145}{\textbf{-8.76\%}} & \textcolor[RGB]{208,77,145}{\textbf{-5.71\%}} \\
\midrule
\rule{0pt}{0.23cm}
GUPNet\cite{lu2021geometry} & 22.76 & 16.46 & 13.72 & 19.90 & 124.57 \\
\rowcolor{cyan!10}
\rule{0pt}{0.23cm}
GUPNet $\oplus$ EH-FAM & \textbf{22.90} & \textbf{16.51} & \textbf{13.92} & \textbf{18.16} & \textbf{115.53} \\
\rule{0pt}{0.23cm}
\emph{Improvement} & \textcolor[RGB]{208,77,145}{\textbf{+0.14}} & \textcolor[RGB]{208,77,145}{\textbf{+0.05}} & \textcolor[RGB]{208,77,145}{\textbf{+0.20}} & \textcolor[RGB]{208,77,145}{\textbf{-8.74\%}} & \textcolor[RGB]{208,77,145}{\textbf{-7.25\%}} \\
\midrule
\rule{0pt}{0.23cm}
MonoLSS\cite{li2024monolss} & 25.91 & 18.29 & 15.94 & 20.51 & 126.13 \\
\rowcolor{cyan!10}
\rule{0pt}{0.23cm}
MonoLSS $\oplus$ EH-FAM & \textbf{26.62} & \textbf{19.30} & \textbf{16.09} & \textbf{18.77} & \textbf{117.09} \\
\rule{0pt}{0.23cm}
\emph{Improvement} & \textcolor[RGB]{208,77,145}{\textbf{+0.71}} & \textcolor[RGB]{208,77,145}{\textbf{+1.01}} & \textcolor[RGB]{208,77,145}{\textbf{+0.15}} & \textcolor[RGB]{208,77,145}{\textbf{-8.48\%}} & \textcolor[RGB]{208,77,145}{\textbf{-7.17\%}} \\
\bottomrule[0.15em]
\end{tabular}
\end{center}
\vspace{-0.25cm}
\label{ablation EH-FAM}
\end{table}

\emph{3) Selective Confidence-Guided Heatmap Loss:} The Selective Confidence-Guided Heatmap Loss guides the network to focus more on high-confidence samples. 
As shown in Table \ref{ablation total}, when used alone, this loss achieves the largest improvement in the baseline model’s $A{P_{3D|R40}}$ for the Car category at the \emph{Easy} level, reaching 25.87\%. When using the other two modules, adding the loss further improves the $A{P_{3D|R40}}$ by 0.37\%. This demonstrates that the loss function is particularly beneficial for detecting \emph{Easy} samples in the KITTI dataset. The likely reason is that the loss encourages the model to rely more on high-confidence detections, and \emph{Easy} samples are generally easier to detect, leading to higher confidence scores.

\begin{figure*}
    \centering
    \subfloat[ MonoLSS]{
        \centering
        \includegraphics[width=0.3264\linewidth]{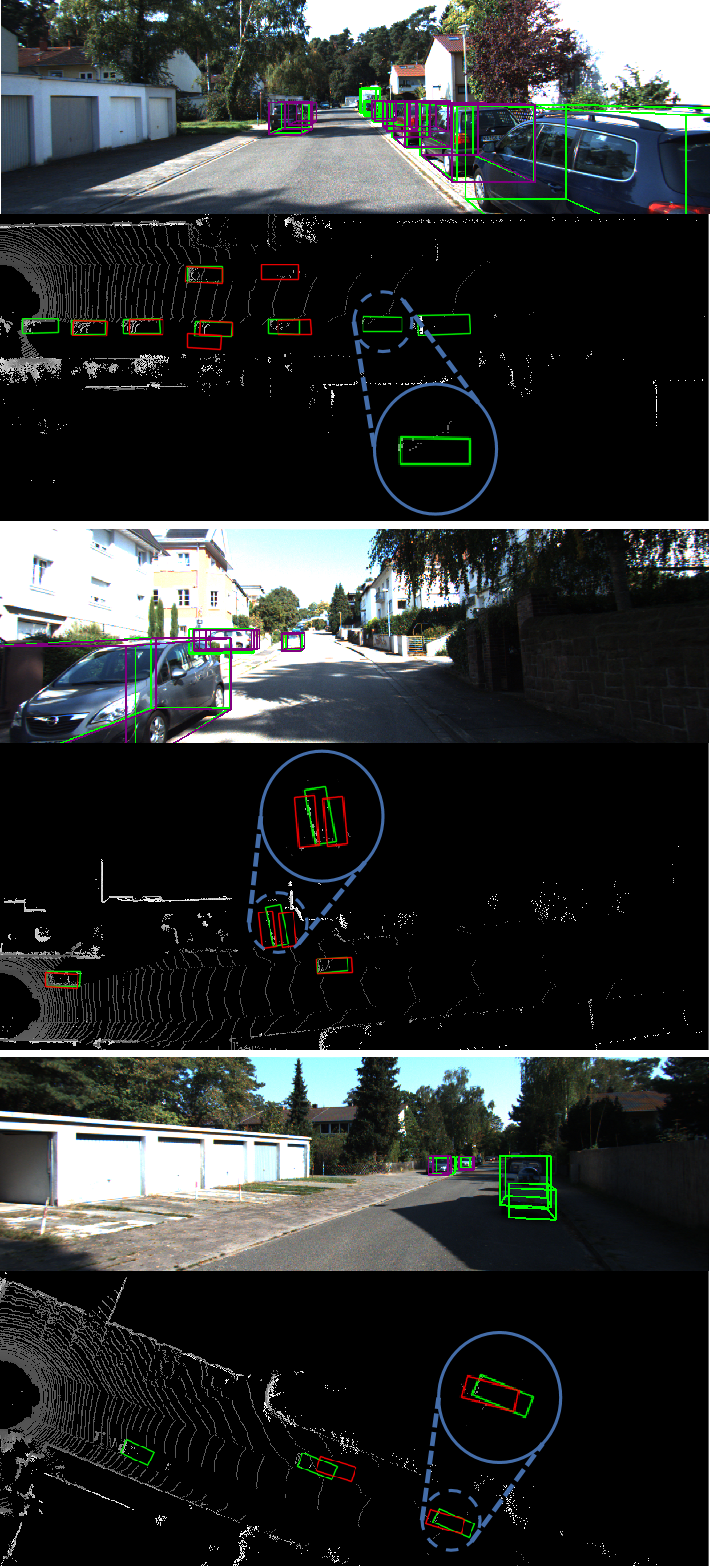}%
        \label{fig:visa}%
    }\hfill
    \subfloat[ MonoDETR]{
        \centering
        \includegraphics[width=0.3264\linewidth]{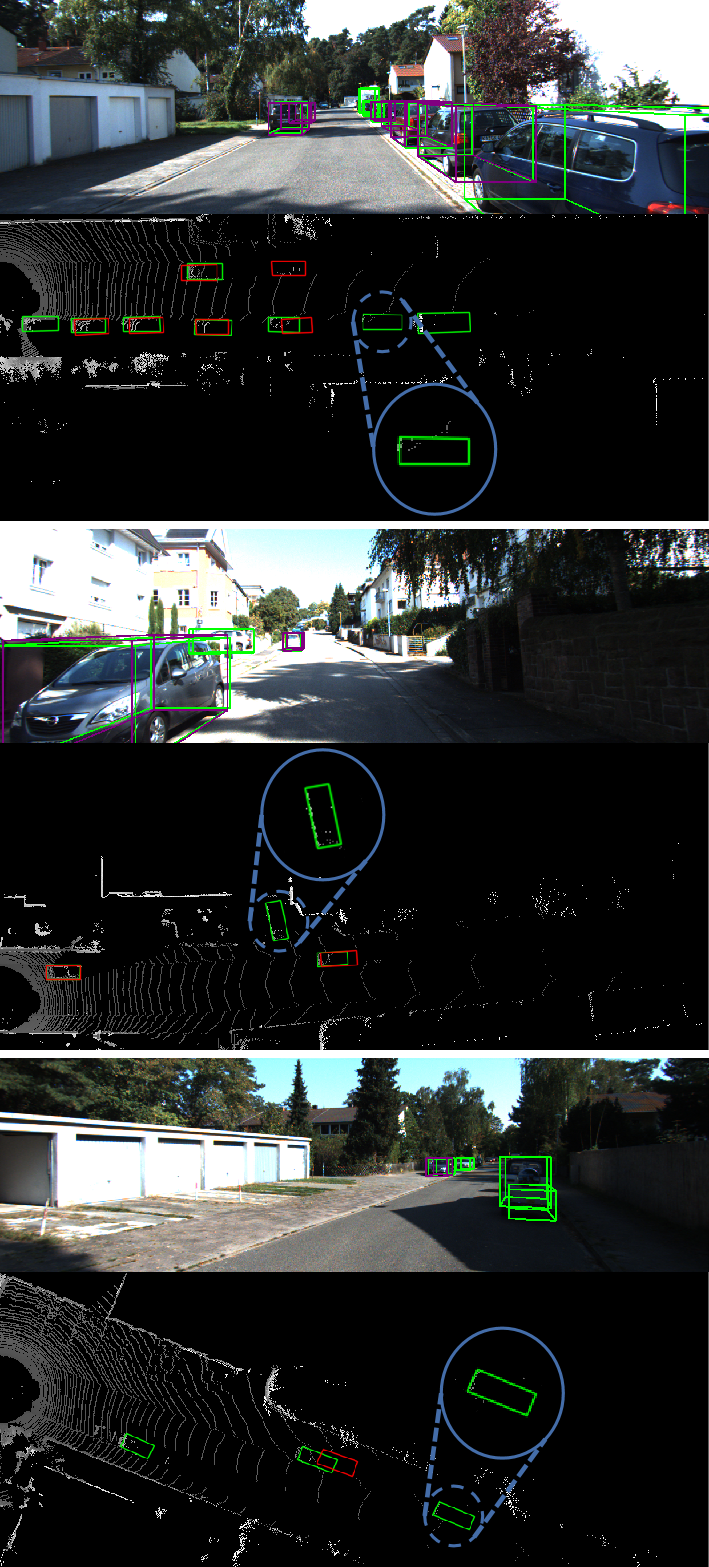}%
        \label{fig:visb}%
    }\hfill
    \subfloat[ MonoASRH]{
        \centering
        \includegraphics[width=0.3264\linewidth]{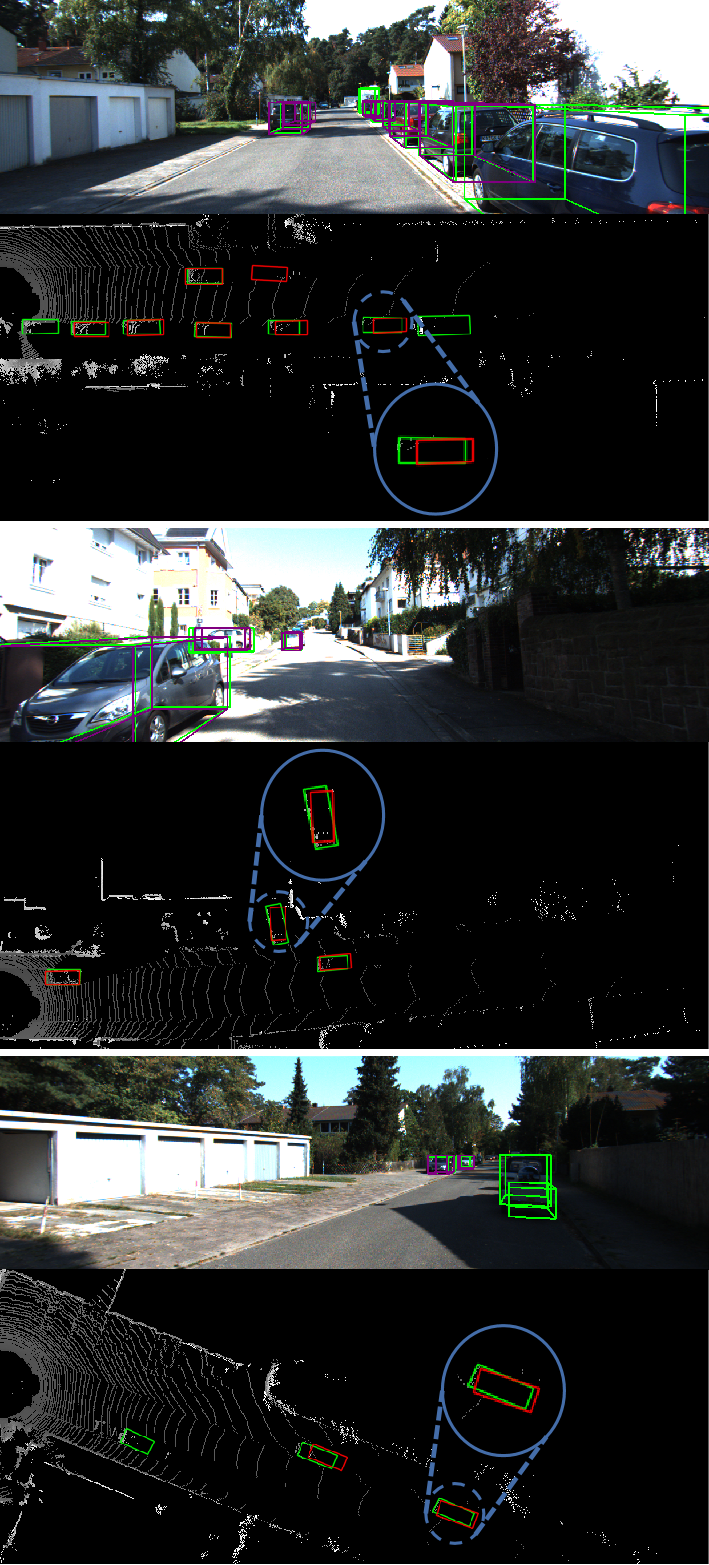}%
        \label{fig:visc}%
    }
    \caption{Qualitative comparison results of 3D object detection on the KITTI val set: (a) MonoLSS\cite{li2024monolss}, (b) MonoDETR\cite{zhang2023monodetr}, and (c) Ours. In each group of images, the first row shows the camera view, and the second row shows the BEV view. In the camera view, predicted and ground truth 3D bounding boxes are shown in purple and green, respectively. In the BEV view, their projections are represented in red and green.}
    \label{fig:vis}
    \vspace{-0.3cm}
\end{figure*}

\subsection{Visualization}
In Fig. \ref{fig:vis}, we compare the proposed MonoASRH with the Transformer-based MonoDETR\cite{zhang2023monodetr} and center-based MonoLSS\cite{li2024monolss}. Each group of images shows the camera view and BEV view. Compared to the other methods, MonoASRH produces higher-quality 3D bounding boxes across various scenes. Notably, it excels at detecting and localizing small distant objects, such as the white vehicle on the right in the third group of images, demonstrating the benefit of incorporating scale information. MonoASRH also provides more accurate results for fully visible objects. Additionally, it more effectively detects partially occluded objects, as shown on the left in the second group of images.

\begin{figure}
    \centering
    \includegraphics[width=0.9\linewidth]{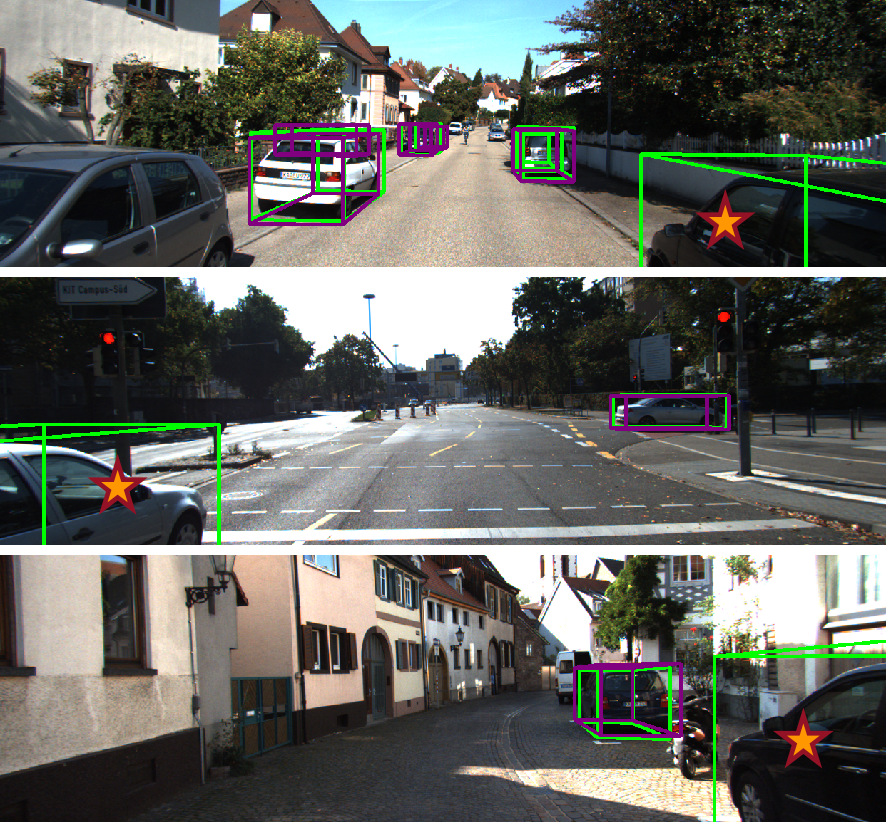}%
    \caption{Qualitative results of failure cases on the KITTI val set. Predicted and ground truth 3D bounding boxes are shown in purple and green, respectively. The pentagrams represent objects that were not detected by the MonoASRH but were annotated in the ground truth.}
    \label{fig:Truncated}
    \vspace{-0.35cm}
\end{figure}

\section{Limitation}
While MonoASRH achieves state-of-the-art performance, it has several limitations. As Fig. \ref{fig:Truncated} illustrates, MonoASRH may underperform when detecting heavily truncated objects (e.g., objects partially outside the image frame). We hypothesize that the aforementioned limitations may arise from the ASRH module's reliance on 2D scale information extraction. Truncation often results in incomplete 2D bounding boxes, leading to inaccurate scale feature extraction by ASRH. This issue is particularly pronounced in crowded urban scenarios where truncation is common. Additionally, ASRH's performance depends mainly on the accuracy of initial 2D object center predictions. Truncated objects often cause shifts in predicted center points, and errors in 2D detection (e.g., missed or mislocalized centers) propagate to the 3D regression stage, degrading overall performance.

Furthermore, while MonoASRH focuses on monocular 3D object detection, its potential in multi-view 3D detection tasks remains unexplored. By effectively bridging 2D and 3D detection, MonoASRH may demonstrate promising adaptability for multi-view 3D detection scenarios. The above limitations underscore future research directions, such as integrating uncertainty estimation for truncated objects, designing robust 2D-3D joint training strategies, and exploring extensions to multi-view 3D detection tasks.

\section{Conclusion}
We propose MonoASRH, which introduces Efficient Hybrid Feature Aggregation Module (EH-FAM) and Scale-Aware 3D Regression Head (ASRH) for monocular 3D detection. Our plug-and-play EH-FAM employs a hybrid architecture to efficiently aggregate features across different scales, endowing the model with richer scale-aware semantic information. ASRH dynamically adjusts the network’s receptive field according to the scale of objects, thereby enhancing 3D detection accuracy. Experimental results on the KITTI and Waymo benchmark demonstrate that MonoASRH achieves state-of-the-art performance. 
However, MonoASRH may underperform when detecting truncated objects, as these objects often lead to inaccuracies in the acquisition of scale features by ASRH. Future research will focus on overcoming this challenge.

\nocite{*}
	\bibliography{bibtex/bib/ref.bib}
	\bibliographystyle{IEEEtran}

\newpage

 




\vfill

\end{document}